\documentclass{article} 
\usepackage{echoDino,times}
\setcitestyle{numbers,square,comma}

\usepackage{hyperref}
\hypersetup{hidelinks}
\usepackage{url}

\usepackage{graphicx}
\usepackage{natbib}
\usepackage{multirow}
\usepackage{arydshln}

\usepackage{subcaption}

\usepackage[table]{xcolor}
\usepackage{booktabs}
\usepackage{longtable}
\usepackage{pdflscape}
\usepackage{array}
\usepackage{tabularx}

\title{EchoDino: A pediatric foundation model for transferable echocardiographic analysis across the lifespan}

\author{%
Sheng~Cheng\textsuperscript{1},
Donnchadh M.~O'Sullivan\textsuperscript{2,3},
Daniel~J.~Penny\textsuperscript{2,3},
Craig~G. Rusin\textsuperscript{2,3},
Minh~B.~Nguyen\textsuperscript{2,3,*}, and
Devika~Subramanian\textsuperscript{1,*,\dag }\\[0.8em]
{\small
\textsuperscript{1}Department of Computer Science, Rice University, Houston, TX, USA\\
\textsuperscript{2}Division of Pediatric Cardiology, Baylor College of Medicine, Houston, TX, USA\\
\textsuperscript{3}Texas Children's Hospital, Houston, TX, USA\\
\textsuperscript{*}Co-senior authors
\textsuperscript{\dag}Correspondence: devika@rice.edu}
}

\iclrfinalcopy 
\newcounter{suppnote}
\renewcommand{\thesuppnote}{\arabic{suppnote}}
\newcommand{\suppnote}[2]{%
  \refstepcounter{suppnote}%
  \section*{Supplementary Note \thesuppnote\quad #1}%
  \label{#2}%
}

\begin{document}

\maketitle
\begin{abstract}
Echocardiography is the most widely used cardiac imaging modality, yet interpretation demands integrating visual evidence across global anatomy, localized structures and dynamic cardiac motion. Machine-learning models have automated individual tasks, but they are typically built for a single purpose and depend on expensively labeled datasets—a barrier particularly acute in pediatric care, where data are scarce and anatomy changes with age. Here we present EchoDino, a self-supervised foundation model for echocardiography, created by adapting the DINOv3 framework to 3.7 million frames from 1.7 million unlabeled pediatric echocardiography videos. With its encoder frozen, EchoDino produces representations that capture global context, local anatomy, and dense spatial detail. We introduce Motion-biased Entropy Maximization Sampling (MEMS) to select the most informative frames for video-level analysis. Across nine pediatric and adult datasets, EchoDino outperformed strong baseline models, raising view-classification accuracy from 0.609 to 0.889 and the area under the receiver operating characteristic curve for structural-heart-disease detection from 0.811 to 0.872, while also cutting age-estimation error from 3.857 to 1.389 years, achieving the best segmentation accuracy and lowering ejection-fraction errors. By generalizing from label-free pediatric data to adult echocardiography, EchoDino offers a versatile foundation for cardiac image analysis across the lifespan. \footnote{Code is available at
\url{https://github.com/Sean-C71/EchoDino}.}
\end{abstract}

\section{Introduction}
Echocardiography is central to cardiovascular care, offering a portable, real-time window onto cardiac structure and function. That versatility also makes echocardiographic data unusually heterogeneous. A single comprehensive study can combine two-dimensional, multi-cycle cine loops, still frames, color Doppler, M-mode, and sweeps acquired across multiple probe positions and imaging conditions~\cite{EchoBase1,EchoBase2}. The evidence that downstream tasks draw on is just as varied. View classification depends on recognizing the whole field of view; chamber quantification and segmentation demand precise localization of endocardial and myocardial boundaries; and functional or diagnostic assessment often hinges on how those structures move through the cardiac cycle~\citep{EchoDyanmic,EchoPed,CAMUS,echoNetLVH}. Because the heart is a moving, three-dimensional organ imaged across multiple views and modes, a useful representation must capture spatial detail, temporal dynamics, and multimodal context simultaneously. Nowhere is this harder than in pediatric echocardiography, where cardiac structure and function change continuously with growth and the definition of normal shifts across development.

Most current systems manage this complexity by narrowing their scope. Task-specific models are trained for a single view, structure, or clinical endpoint~\citep{EchoDyanmic,EchoPed,CAMUS,echoNetLVH}, while recent generalist models consolidate several predictions through paired reports or predefined supervision~\citep{EchoCLIP,EchoPrime,PanEcho}. These approaches have markedly advanced automated interpretation, but the representations they learn are inherently bounded by the annotations, reporting language, and output scales seen during training. Curating the large labeled datasets they require is prohibitively labor-intensive. Supervised objectives also constrain what a model encodes because they reward features that minimize a fixed clinical loss while discarding subtle anatomical and physiological variation.

Self-supervised learning offers a way past this bottleneck, and early echocardiographic models have exploited it using large collections of individual images or explicit spatiotemporal pretraining~\citep{EchoApex,EchoFM}. It remains unknown, however, whether a single label-free representation pretrained solely on pediatric echocardiograms can simultaneously retain global context, fine-grained anatomy, and temporal information while still transferring to adult examinations. Existing methods often depend on heavy downstream fine-tuning or sample the uneven temporal structure of the cardiac cycle poorly. What is needed instead is a general-purpose foundation model with a frozen encoder whose representations support spatial and temporal tasks across patient ages through lightweight task-specific readouts.

\begin{figure}[!htbp]
    \centering
    \includegraphics[width=0.99\textwidth]{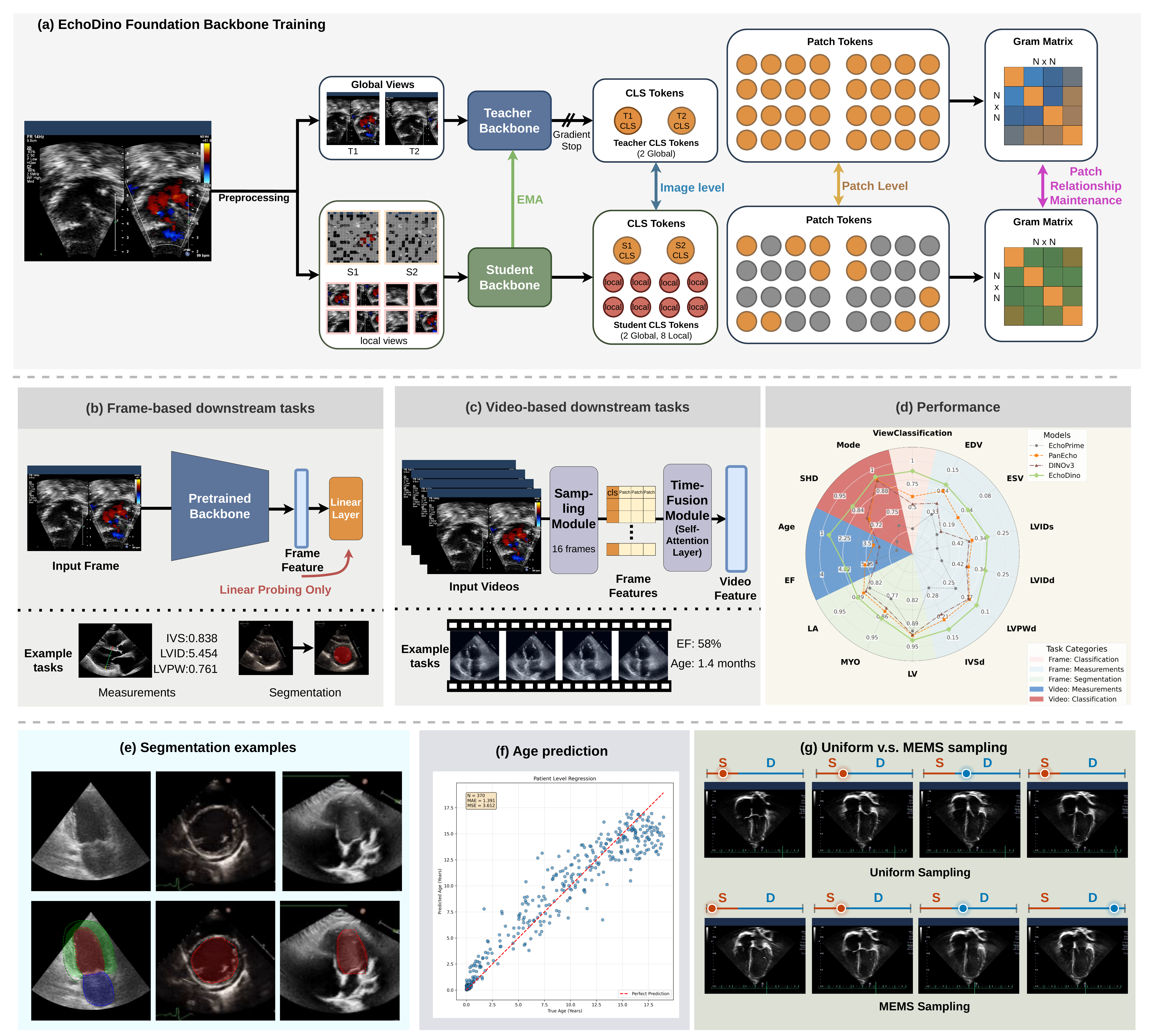}
    \caption{\textbf{Overview of EchoDino.} (a) EchoDino adapts a DINOv3 ViT-L backbone using 3,747,848 unlabeled pediatric echocardiography frames. Self-supervised training aligns image-level representations across augmented views to capture global acquisition context, predicts teacher features for masked patches to retain localized anatomy, and preserves dense relationships among patch tokens during extended training. Training continued for 300,000 steps, and the final checkpoint was used for all downstream evaluations. (b) Frame-based tasks use EchoDino's global class tokens, local patch tokens, or multiscale intermediate features with a frozen encoder. (c) For video-based tasks, a lightweight temporal module aggregates frozen frame representations to produce video-level predictions. (d) EchoDino consistently outperforms baseline models across frame- and video-based tasks. (e) EchoDino accurately performs segmentation outputs across views. Dashed contours indicate ground truth, and filled regions show model predictions. (f) EchoDino predicts age across pediatric age groups in an expected non-linear fashion. (g) Uniform and MEMS frame selections from the same video. The progress bar above each frame indicates its relative position between systole (S) and diastole (D). Compared with uniform frame sampling, the proposed Motion-biased Entropy Maximization Sampling prioritizes feature-diverse frames with substantial motion.}
    \label{fig:EchoDino_arch}
\end{figure}

Here we present EchoDino, a self-supervised foundation model for pediatric echocardiography built on the DINOv3 framework~\citep{DinoV3,Gram1,Gram2}. DINOv3 learns global image context and local patch-level features from images alone, without task labels or paired clinical text, which suits echocardiography's multiscale visual structure. Starting from an encoder pretrained on billions of natural images, we continued self-supervised training on 3.7 million unlabeled frames drawn from 1.7 million pediatric echocardiography videos. The encoder remained frozen, while lightweight task-specific readouts were trained for downstream evaluation. The frozen encoder exposes three complementary feature types: a class token that summarizes the whole frame, patch tokens that retain spatially resolved local information, and intermediate features from earlier layers that preserve the boundary detail needed for segmentation (Fig.~\ref{fig:EchoDino_arch}a).

On this shared representation, we build two lightweight extensions that route its spatial and temporal information to the task at hand, rather than acting as separate models. For frame-level prediction, vision transformers conventionally rely on the global class token alone, discarding the spatial geometry required for precise localization and measurement; EchoDino-PATCH recombines the class token with spatially resolved patch tokens to restore that local detail. For video-level prediction, fixed compute budgets are usually met by sampling frames at uniform intervals, a simple but content-agnostic strategy that is ill-suited to the periodic cardiac cycle~\citep{Q-frame}. Because systole and diastole are unequal in length and their balance shifts with heart rate—an effect especially pronounced at the fast, variable rates of children—uniformly spaced frames cover the cycle unevenly and can miss brief but decisive states such as end-systole. Motion-biased Entropy Maximization Sampling (MEMS) instead selects high-motion, feature-diverse frames while using temporal anchors to preserve whole-video coverage. A lightweight transformer then fuses the selected frames while preserving their spatial positions and source-frame timing (Fig.~\ref{fig:EchoDino_arch}b,c).

We evaluate EchoDino across five internal pediatric cohorts, one external pediatric cohort, and three external adult datasets spanning view classification, cardiac measurement, chamber segmentation, age estimation, ejection-fraction (EF) regression, structural heart disease (SHD) detection, and sweep-mode recognition. Although its echocardiography-specific adaptation used pediatric videos alone, EchoDino achieved the best performance among the tested methods on the adult benchmarks for ventricular measurement, chamber segmentation, and ejection-fraction prediction. Under a strictly frozen-encoder protocol, it consistently surpassed the unadapted DINOv3 encoder and generally exceeded the supervised PanEcho and EchoPrime foundation models. EchoDino-PATCH was strongest on localized anatomical targets, while MEMS reduced ejection-fraction error relative to uniform sampling across three independent datasets and retained robust sweep recognition. Together, these results show that a single label-free echocardiography foundation model can support global, localized, dense, and temporally fused predictions across pediatric and adult populations (Fig.~\ref{fig:EchoDino_arch}).

\section{Results}

\subsection{Datasets and experimental setup}

\begin{table}[htbp]
\centering
\caption{\textbf{Cohorts used for EchoDino pretraining and downstream evaluation.} TCH denotes Texas Children's Hospital, and all TCH-prefixed cohorts are internal pediatric datasets. TCH-SHD is a subset of TCH-Complex, consisting of 56.8\% DICOM files and 57.4\% studies. The remaining TCH downstream cohorts contained no studies included in TCH-Complex. Input counts follow the native unit used to define each cohort: pretraining frames for TCH-Complex, retained frames for TCH-Age and CAMUS, and retained videos for the remaining evaluation cohorts. F and V denote frame- and video-level tasks, respectively; EF, ejection fraction; LA, left atrium; LV, left ventricle; SHD, structural heart disease. Detailed cohort definitions and target distributions are provided in Sec.~\ref{sec:datasets}.}
\label{tab:tch_cohort_summary}
\small
\setlength{\tabcolsep}{3pt}
\renewcommand{\arraystretch}{1.08}
\begin{tabularx}{\textwidth}{@{}
    >{\raggedright\arraybackslash}p{2.5cm}
    l
    l
    >{\raggedright\arraybackslash}p{1.4cm}
    >{\raggedright\arraybackslash}X@{}}
\toprule
\textbf{Dataset} & \textbf{Patients} & \textbf{Studies} & \textbf{Inputs} & \textbf{Task(s)} \\
\midrule
\multicolumn{5}{@{}l}{\textit{Internal pediatric pretraining}} \\
\addlinespace[1.5pt]
TCH-Complex & 13,323 & 17,984 & 3,747,848 & Self-supervised pretraining [F] \\
\addlinespace[4pt]
\midrule
\multicolumn{5}{@{}l}{\textit{Internal pediatric evaluation}} \\
\addlinespace[1.5pt]
TCH-SHD & 12,517 & 16,427 & 1,015,795 & Binary SHD classification [V] \\
TCH-View & 298 & 301 & 10,522 & 52-class view classification [F] \\
TCH-EF & 2,637 & 3,864 & 13,809 & EF regression [V] \\
TCH-Age & 2,915 & 3,865 & 101,636 & Age regression and interval classification [F] \\
TCH-Sweep & 313 & 316 & 13,410 & Four-class acquisition-mode classification [V] \\
\addlinespace[4pt]
\midrule
\multicolumn{5}{@{}l}{\textit{External pediatric evaluation}} \\
\addlinespace[1.5pt]
EchoNet-Pediatric & 1,958 & 4,467 & 7,643 & LV segmentation [F]; EF regression [V] \\
\addlinespace[4pt]
\midrule
\multicolumn{5}{@{}l}{\textit{External adult evaluation}} \\
\addlinespace[1.5pt]
CAMUS & 500 & 500 & 2,000 & LV, myocardium, and LA segmentation [F] \\
EchoNet-LVH & 12,000 & 12,000 & 12,000 & LV dimension and wall-thickness measurements [F] \\
EchoNet-Dynamic & 10,030 & 10,030 & 10,030 & LV volume estimation and segmentation [F]; EF regression [V] \\
\bottomrule
\end{tabularx}
\end{table}

We organized our evaluation around the distinct types of information required from an echocardiogram. Frame-level experiments assessed acquisition context, localized cardiac geometry, anatomical boundaries, and age-related developmental information. Video-level experiments assessed cardiac function and structural heart disease (SHD) classification after temporal fusion. Comparisons followed a consistent hierarchy: EchoDino assessed baseline feature extraction, EchoDino-PATCH assessed localized spatial reaggregation, and EchoDino-MEMS assessed temporal evidence selection. To rigorously evaluate the learned representations, the foundation model backbone itself remained strictly frozen during downstream evaluation, with only the task-specific heads being trainable. Public external datasets were evaluated using their official partitions when available. For TCH cohorts lacking predefined partitions, we created patient-disjoint, label-stratified training, validation, and test sets; cohort-specific partition sizes and scanner-vendor coverage are reported in Supplementary Tables~\ref{si-tab:tch_view_distribution}--\ref{si-tab:tch_vendor_distribution}. Hyperparameters and model checkpoints were selected using validation performance only, and final performance was reported on the held-out test sets. Unless otherwise specified, performance is reported as bootstrap medians with percentile 95\% confidence intervals, and two-sided \(P\) values reflect paired model comparisons; full statistical procedures are described in Methods.

EchoDino was initialized from natural-image DINOv3 weights and subsequently adapted using only pediatric echocardiography videos. We therefore evaluated its transferability across five internal pediatric datasets, one external pediatric dataset, and three external adult datasets (Table~\ref{tab:tch_cohort_summary} and Sec.~\ref{sec:datasets}). Fig.~\ref{fig:EchoDino_arch}d summarizes performance across classification, measurement, segmentation, and video-prediction tasks. These results demonstrate that EchoDino generally outperforms the evaluated baseline encoders under a frozen-encoder protocol across pediatric and adult cohorts on diverse downstream tasks, including global (view classification, age prediction), localized (cardiac measurement, structural heart disease diagnosis), dense (chamber segmentation), and temporal endpoints (ejection-fraction regression, sweep-mode recognition).

\subsection{EchoDino captures global acquisition context and localized cardiac anatomy}

\textbf{EchoDino captures the global organization of echocardiographic images.} View classification tests whether a frame representation contains sufficient acquisition context to distinguish clinically recognized echocardiographic acquisitions. On the TCH-View dataset, this task comprised 52 labels that jointly encoded acquisition mode (e.g., two-dimensional imaging or color Doppler), standard view family, and a reference designation, anatomical level, or focus (e.g., a parasternal long-axis reference view or a parasternal short-axis view at the mitral-valve/papillary-muscle level). We evaluated the frozen representation using K-nearest-neighbor  (KNN) classification and a linear probe. EchoDino achieved an accuracy of 0.850 at k = 10 and a linear-probe accuracy of 0.889 (Table~\ref{tab:view_classification}). By comparison, PanEcho achieved 0.504 and 0.609, whereas DINOv3 achieved 0.369 and 0.464, respectively. All comparisons with EchoDino were highly significant (p $<$ 0.001), and its k-nearest-neighbor accuracy remained robust at 0.848 for both k = 20 and k = 30. Detailed per-class results are provided in Supplementary Note~\ref{si-note:view-classification} and Supplementary Table~\ref{si-tab:view_classification_per_class}. This task-head-free performance establishes that EchoDino’s frozen representation directly and accurately encodes global echocardiographic acquisition context.

    \begin{table}[!htbp]
        \caption{\textbf{View-classification accuracy on the TCH-View dataset.} Values are bootstrap medians (95\% CIs). P values compare each baseline with EchoDino.}
        \label{tab:view_classification}
        \centering
        \small
        \setlength{\tabcolsep}{4pt}
        \begin{tabular}{lcccc}
        \toprule
        \multirow[c]{2}{*}{\bf Method}
            & \multicolumn{3}{c}{\bf KNN}
            & \multirow[c]{2}{*}{\bf Linear probe} \\
        \cmidrule(lr){2-4}
            & \bf $k=10$ & \bf $k=20$ & \bf $k=30$ & \\
        \midrule
        \raisebox{0.8ex}{EchoPrime}
            & \shortstack{0.224 (0.206--0.244)\\{\scriptsize [$p<0.001$]}}
            & \shortstack{0.229 (0.210--0.248)\\{\scriptsize [$p<0.001$]}}
            & \shortstack{0.228 (0.210--0.248)\\{\scriptsize [$p<0.001$]}}
            & \shortstack{0.274 (0.246--0.302)\\{\scriptsize [$p<0.001$]}} \\
        \raisebox{0.8ex}{PanEcho}
            & \shortstack{0.504 (0.482--0.528)\\{\scriptsize [$p<0.001$]}}
            & \shortstack{0.496 (0.475--0.518)\\{\scriptsize [$p<0.001$]}}
            & \shortstack{0.483 (0.462--0.506)\\{\scriptsize [$p<0.001$]}}
            & \shortstack{0.609 (0.579--0.640)\\{\scriptsize [$p<0.001$]}} \\
        \midrule
        \raisebox{0.8ex}{DINOv3}
            & \shortstack{0.369 (0.347--0.389)\\{\scriptsize [$p<0.001$]}}
            & \shortstack{0.366 (0.345--0.387)\\{\scriptsize [$p<0.001$]}}
            & \shortstack{0.359 (0.337--0.380)\\{\scriptsize [$p<0.001$]}}
            & \shortstack{0.464 (0.433--0.495)\\{\scriptsize [$p<0.001$]}} \\
        \midrule
        EchoDino
            & \bf 0.850 (0.834--0.866)
            & \bf 0.848 (0.833--0.864)
            & \bf 0.848 (0.833--0.865)
            & \bf 0.889 (0.868--0.908) \\
        \bottomrule
        \end{tabular}
    \end{table}

    \textbf{EchoDino preserves localized cardiac anatomy beyond its global image representation.} 
    Left ventricular measurement tests whether frozen features successfully encode chamber size, ventricular volume, and myocardial wall thickness at specific cardiac phases. We evaluated end-diastolic volume (EDV) and end-systolic volume (ESV) on the EchoNet-Dynamic dataset, alongside four linear chamber and wall measurements on the EchoNet-LVH dataset. Using only the global class token, EchoDino produced lower errors than DINOv3 using the same class-token readout for all six endpoints, yielding relative reductions of 12.6–20.1\% (Table~\ref{tab:lv_measurement_unified}). However, the EchoDino class-token readout remained behind PanEcho, indicating that a single global token does not fully expose the necessary local geometry. Conversely, EchoDino-PATCH (which reaggregates this global token with spatial patch tokens) achieved the lowest mean absolute error (MAE) for every endpoint. Relative to the EchoDino class-token readout, EchoDino-PATCH reduced EDV and ESV errors by 28.0\% and 41.9\%, respectively, and reduced the four linear-measurement errors by 23.6–28.9\%. EchoDino-PATCH also consistently outperformed DINOv3-PATCH across all six endpoints. This comparison isolates two complementary strengths: EchoDino intrinsically preserves local ventricular structure, and the PATCH mechanism successfully makes that information accessible for precise cardiac measurement.

    \begin{table}[!htbp]
        \caption{\textbf{Left ventricular (LV) measurement performance.} Values are bootstrap medians (95\% CIs), and all metrics are mean absolute errors (MAE; $\downarrow$). EDV and ESV are reported in normalized 100-mL units (for example, 0.180 corresponds to 18.0\,mL); LVIDs, LVIDd, LVPWd, and IVSd are reported in centimetres. Where reported, P values compare each method with EchoDino-PATCH.}
        \label{tab:lv_measurement_unified}
        \centering
        \small
        \setlength{\tabcolsep}{3pt}
        \resizebox{\textwidth}{!}{%
        \begin{tabular}{lcccccc}
        \toprule
         & \multicolumn{2}{c}{\bf EchoNet-Dynamic} & \multicolumn{4}{c}{\bf EchoNet-LVH} \\
        \cmidrule(lr){2-3}\cmidrule(lr){4-7}
        \bf Method & \bf EDV$\downarrow$ & \bf ESV$\downarrow$ & \bf LVIDs$\downarrow$ & \bf LVIDd$\downarrow$ & \bf LVPWd$\downarrow$ & \bf IVSd$\downarrow$ \\
        \midrule
        \raisebox{0.8ex}{EchoPrime}
            & \shortstack{0.333 (0.314--0.355)\\{\scriptsize [$p<0.001$]}}
            & \shortstack{0.225 (0.212--0.239)\\{\scriptsize [$p<0.001$]}}
            & \shortstack{0.505 (0.463--0.552)\\{\scriptsize [$p<0.001$]}}
            & \shortstack{0.488 (0.445--0.532)\\{\scriptsize [$p<0.001$]}}
            & \shortstack{0.222 (0.201--0.243)\\{\scriptsize [$p<0.001$]}}
            & \shortstack{0.299 (0.265--0.335)\\{\scriptsize [$p<0.001$]}} \\
        \addlinespace[1.5pt]
        \raisebox{0.8ex}{PanEcho}
            & \shortstack{0.236 (0.224--0.250)\\{\scriptsize [$p<0.001$]}}
            & \shortstack{0.162 (0.153--0.171)\\{\scriptsize [$p<0.001$]}}
            & \shortstack{0.369 (0.339--0.401)\\{\scriptsize [$p<0.001$]}}
            & \shortstack{0.383 (0.349--0.415)\\{\scriptsize [$p<0.001$]}}
            & \shortstack{0.165 (0.149--0.179)\\{\scriptsize [$p<0.001$]}}
            & \shortstack{0.204 (0.180--0.230)\\{\scriptsize [$p<0.001$]}} \\
        \midrule
        \raisebox{0.8ex}{DINOv3 (class token)}
            & \shortstack{0.286 (0.271--0.304)\\{\scriptsize [$p<0.001$]}}
            & \shortstack{0.214 (0.203--0.227)\\{\scriptsize [$p<0.001$]}}
            & \shortstack{0.466 (0.423--0.516)\\{\scriptsize [$p<0.001$]}}
            & \shortstack{0.482 (0.441--0.526)\\{\scriptsize [$p<0.001$]}}
            & \shortstack{0.218 (0.197--0.239)\\{\scriptsize [$p<0.001$]}}
            & \shortstack{0.294 (0.262--0.329)\\{\scriptsize [$p<0.001$]}} \\
        \addlinespace[1.5pt]
        \raisebox{0.8ex}{DINOv3-PATCH}
            & \shortstack{0.203 (0.193--0.215)\\{\scriptsize [$p<0.001$]}}
            & \shortstack{0.112 (0.104--0.119)\\{\scriptsize [$p=0.006$]}}
            & \shortstack{0.382 (0.347--0.417)\\{\scriptsize [$p<0.001$]}}
            & \shortstack{0.391 (0.356--0.426)\\{\scriptsize [$p<0.001$]}}
            & \shortstack{0.168 (0.152--0.183)\\{\scriptsize [$p<0.001$]}}
            & \shortstack{0.221 (0.196--0.247)\\{\scriptsize [$p<0.001$]}} \\
        \midrule
        \raisebox{0.8ex}{EchoDino (class token)}
            & \shortstack{0.250 (0.237--0.264)\\{\scriptsize [$p<0.001$]}}
            & \shortstack{0.179 (0.170--0.189)\\{\scriptsize [$p<0.001$]}}
            & \shortstack{0.406 (0.374--0.442)\\{\scriptsize [$p<0.001$]}}
            & \shortstack{0.421 (0.385--0.456)\\{\scriptsize [$p<0.001$]}}
            & \shortstack{0.190 (0.173--0.207)\\{\scriptsize [$p<0.001$]}}
            & \shortstack{0.235 (0.209--0.262)\\{\scriptsize [$p<0.001$]}} \\
        \addlinespace[1.5pt]
        EchoDino-PATCH
            & \bf 0.180 (0.171--0.190)
            & \bf 0.104 (0.097--0.110)
            & \bf 0.310 (0.284--0.338)
            & \bf 0.317 (0.290--0.348)
            & \bf 0.135 (0.122--0.148)
            & \bf 0.175 (0.156--0.195) \\
        \bottomrule
        \end{tabular}%
        }
    \end{table}

    \textbf{EchoDino retains fine anatomical boundaries across echocardiographic domains.} 
    Dense segmentation tests whether frozen features support pixel-level delineation of cardiac chambers and the myocardium. We evaluated left ventricular (LV) segmentation on EchoNet-Dynamic and EchoNet-Pediatric, as well as multistructure segmentation on CAMUS, using a one-layer linear segmentation head for all models. EchoDino significantly exceeded DINOv3 for every structural target (Table~\ref{tab:segmentation_unified}; all $p < 0.001$). LV Dice scores reached 0.910 on EchoNet-Dynamic and 0.903 on EchoNet-Pediatric, exceeding PanEcho by 0.023 and 0.043. On the CAMUS dataset, EchoDino achieved Dice coefficients of 0.932 for the LV, 0.869 for the myocardium, and 0.888 for the left atrium—exceeding PanEcho by 0.013, 0.029, and 0.020, respectively. Representative examples (Fig.~\ref{fig:EchoDino_arch}e) show predicted chamber and myocardial regions accurately aligned with reference contours across apical, short-axis, and long-axis views. These consistent gains across three distinct datasets confirm that EchoDino retains highly accurate boundary-level information for multiple cardiac structures across different age groups.

    \begin{table}[!htbp]
        \caption{\textbf{Segmentation performance across clinical datasets.} Values are bootstrap medians (95\% CIs), and all metrics are Dice coefficients ($\uparrow$). P values compare each baseline with EchoDino.}
        \label{tab:segmentation_unified}
        \centering
        \small
        \setlength{\tabcolsep}{3pt}
        \resizebox{\textwidth}{!}{%
        \begin{tabular}{lccccc}
        \toprule
         & \bf EchoNet-Dynamic & \bf EchoNet-Pediatric & \multicolumn{3}{c}{\bf CAMUS} \\
        \cmidrule(lr){2-2}\cmidrule(lr){3-3}\cmidrule(lr){4-6}
        \bf Method & \bf LV$\uparrow$ & \bf LV$\uparrow$ & \bf LV$\uparrow$ & \bf MYO$\uparrow$ & \bf LA$\uparrow$ \\
        \midrule
        \raisebox{0.8ex}{EchoPrime}
            & \shortstack{0.864 (0.861--0.867)\\{\scriptsize [$p<0.001$]}}
            & \shortstack{0.809 (0.804--0.815)\\{\scriptsize [$p<0.001$]}}
            & \shortstack{0.907 (0.905--0.909)\\{\scriptsize [$p<0.001$]}}
            & \shortstack{0.787 (0.785--0.790)\\{\scriptsize [$p<0.001$]}}
            & \shortstack{0.847 (0.843--0.850)\\{\scriptsize [$p<0.001$]}} \\
        \addlinespace[1.5pt]
        \raisebox{0.8ex}{PanEcho}
            & \shortstack{0.887 (0.884--0.889)\\{\scriptsize [$p<0.001$]}}
            & \shortstack{0.860 (0.855--0.864)\\{\scriptsize [$p<0.001$]}}
            & \shortstack{0.919 (0.917--0.921)\\{\scriptsize [$p<0.001$]}}
            & \shortstack{0.840 (0.838--0.842)\\{\scriptsize [$p<0.001$]}}
            & \shortstack{0.868 (0.864--0.871)\\{\scriptsize [$p<0.001$]}} \\
        \midrule
        \raisebox{0.8ex}{DINOv3}
            & \shortstack{0.888 (0.886--0.890)\\{\scriptsize [$p<0.001$]}}
            & \shortstack{0.864 (0.860--0.868)\\{\scriptsize [$p<0.001$]}}
            & \shortstack{0.917 (0.915--0.918)\\{\scriptsize [$p<0.001$]}}
            & \shortstack{0.826 (0.824--0.828)\\{\scriptsize [$p<0.001$]}}
            & \shortstack{0.860 (0.856--0.863)\\{\scriptsize [$p<0.001$]}} \\
        \midrule
        EchoDino
            & \textbf{0.910 (0.909--0.912)}
            & \textbf{0.903 (0.900--0.906)}
            & \textbf{0.932 (0.930--0.933)}
            & \textbf{0.869 (0.867--0.871)}
            & \textbf{0.888 (0.885--0.891)} \\
        \bottomrule
        \end{tabular}%
        }
    \end{table}

\textbf{EchoDino captures developmental variation in pediatric echocardiograms.} Age prediction tests whether a single frame contains the morphological information associated with cardiac growth from infancy through adolescence. On the TCH-Age dataset, one frame was retained from each source video across a small subset of standard views. We evaluated patient-level age regression and subsequently derived three age classes by thresholding the continuous predictions at 2 and 12 years. Using only the global class token, EchoDino reduced the MAE from 4.178 years with DINOv3 to 2.747 years, outperforming PanEcho (3.857 years) (Table~\ref{tab:age_patient_level}). EchoDino-PATCH further reduced the MAE to 1.389 years, a 49.4\% reduction relative to the EchoDino class-token readout. Fig.~\ref{fig:EchoDino_arch}f plots patient-level predicted age against chronological age in the held-out test set, demonstrating tight agreement across the entire pediatric age spectrum. For interval classification, the EchoDino class-token readout achieved an accuracy of 0.630, compared with 0.397 for DINOv3 and 0.530 for PanEcho. EchoDino-PATCH further increased this accuracy and the corresponding F1 score to 0.884 (all comparisons with EchoDino-PATCH yielded $p < 0.001$). Detailed results and the confusion matrix for this classification task are provided in Supplementary Note~\ref{si-note:age-prediction} and Supplementary Fig.~\ref{si-fig:cm_TCHAge}. The stepwise improvement from DINOv3, through the EchoDino class-token readout, to EchoDino-PATCH highlights increasingly effective access to complex developmental cardiac features.

\begin{table}[htbp]
    \caption{\textbf{Age-prediction performance in years.} Values are bootstrap medians (95\% CIs), and frame-level predictions were aggregated for patient-level evaluation. P values compare each method with EchoDino-PATCH.}
    \label{tab:age_patient_level}
    \centering
    \small
    \setlength{\tabcolsep}{4pt}
    \resizebox{\textwidth}{!}{%
    \begin{tabular}{lcccc}
    \toprule
    \bf Method & \bf MAE $\downarrow$ & \bf MSE $\downarrow$ & \bf Acc $\uparrow$ & \bf F1 $\uparrow$ \\
    \midrule
    \raisebox{0.8ex}{EchoPrime}
        & \shortstack{5.133 (4.854--5.428)\\{\scriptsize [$p<0.001$]}}
        & \shortstack{34.396 (31.585--37.353)\\{\scriptsize [$p<0.001$]}}
        & \shortstack{0.362 (0.316--0.408)\\{\scriptsize [$p<0.001$]}}
        & \shortstack{0.198 (0.155--0.243)\\{\scriptsize [$p<0.001$]}} \\
    \addlinespace[1.5pt]
    \raisebox{0.8ex}{PanEcho}
        & \shortstack{3.857 (3.624--4.089)\\{\scriptsize [$p<0.001$]}}
        & \shortstack{20.005 (17.929--21.970)\\{\scriptsize [$p<0.001$]}}
        & \shortstack{0.530 (0.481--0.578)\\{\scriptsize [$p<0.001$]}}
        & \shortstack{0.461 (0.411--0.511)\\{\scriptsize [$p<0.001$]}} \\
    \midrule
    \raisebox{0.8ex}{DINOv3}
        & \shortstack{4.178 (3.938--4.438)\\{\scriptsize [$p<0.001$]}}
        & \shortstack{23.568 (21.285--26.075)\\{\scriptsize [$p<0.001$]}}
        & \shortstack{0.397 (0.349--0.446)\\{\scriptsize [$p<0.001$]}}
        & \shortstack{0.269 (0.215--0.321)\\{\scriptsize [$p<0.001$]}} \\
    \midrule
    \raisebox{0.8ex}{EchoDino (class token)}
        & \shortstack{2.747 (2.532--2.971)\\{\scriptsize [$p<0.001$]}}
        & \shortstack{11.687 (10.278--13.274)\\{\scriptsize [$p<0.001$]}}
        & \shortstack{0.630 (0.581--0.678)\\{\scriptsize [$p<0.001$]}}
        & \shortstack{0.626 (0.578--0.676)\\{\scriptsize [$p<0.001$]}} \\
    \addlinespace[1.5pt]
    EchoDino-PATCH
        & \textbf{1.389 (1.266--1.524)}
        & \textbf{3.606 (3.051--4.241)}
        & \textbf{0.884 (0.851--0.916)}
        & \textbf{0.884 (0.852--0.916)} \\
    \bottomrule
    \end{tabular}%
    }
\end{table}

\subsection{EchoDino aggregates study-level video evidence for robust structural heart disease diagnosis}

\textbf{EchoDino integrates complementary evidence across heterogeneous echocardiographic acquisitions.} Diagnosing SHD requires integrating evidence of pathologic anatomy and physiology across multiple heterogeneous acquisitions and views. We tested whether a lightweight temporal fusion module could aggregate this distributed evidence to distinguish congenital defects from normal anatomy, all while keeping the underlying frame encoder strictly frozen. On the TCH-SHD dataset, we averaged the logits from every video associated with each study. EchoDino achieved an area under the curve (AUC) of 0.872, outperforming PanEcho (0.811) and DINOv3 (0.776) (Table~\ref{tab:tch_chd_disease}). Furthermore, it achieved an F1 score of 0.778 and an accuracy of 0.788, exceeding the strong baselines significantly ($p<0.001$). These robust gains across both ranking and thresholded metrics confirm EchoDino's ability to aggregate complementary spatial evidence for study-level congenital disease detection.

To understand the granularity of this diagnostic accuracy, we performed a sub-analysis of the specific diagnoses within the SHD-positive class. Detailed phenotype-stratified SHD-positive detection rates are reported in Supplementary Note~\ref{si-note:shd-phenotypes} and Supplementary Table~\ref{si-tab:tch_chd_disease_details}. EchoDino achieved the highest or tied rate for all 24 labels, and the uniquely highest rate for 10. Detection reached a perfect 1.000 for 14 critical labels. For the more prevalent atrial septal defect, ventricular septal defect, and patent ductus arteriosus labels, EchoDino achieved robust detection rates of 0.781, 0.856, and 0.790, respectively. By comparison, the strongest baseline model achieved 0.742, 0.733, and 0.730 on these same anomalies. Ultimately, these results demonstrate that EchoDino leverages highly expressive feature representations to effectively aggregate diagnostic evidence across a highly heterogeneous and clinically relevant spectrum of cardiac anatomy.

\begin{table}[htbp]
    \caption{\textbf{TCH-SHD classification performance.} Values are bootstrap medians (95\% CIs), and study-level predictions were obtained by averaging logits from all videos associated with each study. P values compare each baseline with EchoDino.}
    \label{tab:tch_chd_disease}
    \centering
    \small
    \setlength{\tabcolsep}{3pt}
    \resizebox{\textwidth}{!}{%
    \begin{tabular}{lccccc}
    \toprule
    \bf Method & \bf Accuracy $\uparrow$ & \bf Precision $\uparrow$ & \bf Recall $\uparrow$ & \bf F1 score $\uparrow$ & \bf AUC $\uparrow$ \\
    \midrule
    \raisebox{0.8ex}{EchoPrime}
        & \shortstack{0.621 (0.596--0.644)\\{\scriptsize [$p<0.001$]}}
        & \shortstack{0.612 (0.583--0.639)\\{\scriptsize [$p<0.001$]}}
        & \shortstack{0.582 (0.560--0.603)\\{\scriptsize [$p<0.001$]}}
        & \shortstack{0.571 (0.545--0.595)\\{\scriptsize [$p<0.001$]}}
        & \shortstack{0.644 (0.618--0.669)\\{\scriptsize [$p<0.001$]}} \\
    \addlinespace[1.5pt]
    \raisebox{0.8ex}{PanEcho}
        & \shortstack{0.738 (0.714--0.760)\\{\scriptsize [$p<0.001$]}}
        & \shortstack{0.735 (0.712--0.758)\\{\scriptsize [$p<0.001$]}}
        & \shortstack{0.721 (0.699--0.743)\\{\scriptsize [$p<0.001$]}}
        & \shortstack{0.724 (0.701--0.747)\\{\scriptsize [$p<0.001$]}}
        & \shortstack{0.811 (0.789--0.832)\\{\scriptsize [$p<0.001$]}} \\
    \midrule
    \raisebox{0.8ex}{DINOv3}
        & \shortstack{0.720 (0.696--0.742)\\{\scriptsize [$p<0.001$]}}
        & \shortstack{0.722 (0.698--0.746)\\{\scriptsize [$p<0.001$]}}
        & \shortstack{0.695 (0.673--0.717)\\{\scriptsize [$p<0.001$]}}
        & \shortstack{0.698 (0.674--0.722)\\{\scriptsize [$p<0.001$]}}
        & \shortstack{0.776 (0.751--0.798)\\{\scriptsize [$p<0.001$]}} \\
    \midrule
    EchoDino
        & \textbf{0.788 (0.765--0.806)}
        & \textbf{0.788 (0.766--0.807)}
        & \textbf{0.773 (0.752--0.793)}
        & \textbf{0.778 (0.755--0.797)}
        & \textbf{0.872 (0.852--0.888)} \\
    \bottomrule
    \end{tabular}%
    }
\end{table}

\subsection{MEMS improves EchoDino's phase-dependent functional predictions}

\textbf{EchoDino and MEMS support ejection-fraction prediction across pediatric and adult cohorts.}
Unlike SHD diagnosis, ejection fraction (EF) prediction relies heavily on temporal dynamics, specifically whether the sampled frames adequately isolate key phases of the cardiac cycle, such as end-systole and end-diastole. Although EchoDino was pretrained on pediatric echocardiography, the external adult EchoNet-Dynamic test set provided a direct assessment of transfer across age groups. Using a fixed budget of $K=16$ uniformly sampled frames, EchoDino achieved mean absolute errors (MAEs) of 5.042, 4.800, and 6.410 on the adult EchoNet-Dynamic, EchoNet-Pediatric, and TCH-EF datasets, respectively (Table~\ref{tab:ef_prediction_unified}). These were the lowest MAEs among all uniformly sampled methods across the three cohorts, showing that EchoDino's frozen-frame features provide a strong foundation for temporal fusion in both pediatric and adult echocardiography.

However, while uniform sampling preserves regular temporal spacing, it is content-agnostic (Fig.~\ref{fig:EchoDino_arch}g). Because the relative lengths of systole and diastole shift with heart rate and contractility, uniform spacing can miss brief, critical functional states. We therefore compared this baseline against our Motion-biased Entropy Maximization Sampling (MEMS) strategy, which iteratively selects temporally diverse, high-motion frames while preserving whole-video coverage. Holding the frozen encoder, temporal fusion architecture, and 16-frame budget fixed, MEMS further reduced MAE from 5.042 to 4.674 on the EchoNet-Dynamic dataset, and from 4.800 to 4.555 on the EchoNet-Pediatric dataset. On TCH-EF, MAE decreased from 6.410 to 5.652 (an 11.8\% reduction). The corresponding mean squared error (MSE) reductions were 16.4\%, 9.5\%, and 20.1\%, respectively, with all uniform-to-MEMS comparisons yielding $p\le0.05$. While EchoPrime exhibited the lowest MSE on the EchoNet-Pediatric dataset, the difference with EchoDino-MEMS is not statistically significant at $p=0.226$. Together, these results show that MEMS adds value to EchoDino's temporally expressive features across the pediatric and adult cohorts. Prediction-level comparisons on TCH-EF are provided in Supplementary Note~\ref{si-note:ef-prediction} and Supplementary Fig.~\ref{si-fig:ef-scatter-comparison}.

\begin{table}[htbp]
    \caption{\textbf{Ejection fraction (\%) prediction performance across three clinical datasets.} Values are bootstrap medians (95\% CIs). P values compare each method with EchoDino-MEMS.}
    \label{tab:ef_prediction_unified}
    \centering
    \small
    \setlength{\tabcolsep}{3pt}
    \resizebox{\textwidth}{!}{%
    \begin{tabular}{lcccccc}
    \toprule
     & \multicolumn{2}{c}{\bf EchoNet-Dynamic} & \multicolumn{2}{c}{\bf EchoNet-Pediatric} & \multicolumn{2}{c}{\bf TCH-EF} \\
    \cmidrule(lr){2-3}\cmidrule(lr){4-5}\cmidrule(lr){6-7}
    \bf Method & \bf MAE $\downarrow$ & \bf MSE $\downarrow$ & \bf MAE $\downarrow$ & \bf MSE $\downarrow$ & \bf MAE $\downarrow$ & \bf MSE $\downarrow$ \\
    \midrule
    \raisebox{0.8ex}{EchoPrime}
        & \shortstack{5.291 (5.059--5.561)\\{\scriptsize [$p<0.001$]}}
        & \shortstack{48.433 (43.891--53.973)\\{\scriptsize [$p<0.001$]}}
        & \shortstack{4.801 (4.529--5.064)\\{\scriptsize [$p=0.112$]}}
        & \shortstack{\bf 40.242 (35.541--45.664)\\{\scriptsize [$p=0.226$]}}
        & \shortstack{9.112 (8.799--9.404)\\{\scriptsize [$p<0.001$]}}
        & \shortstack{130.915 (122.732--139.157)\\{\scriptsize [$p<0.001$]}} \\
    \addlinespace[1.5pt]
    \raisebox{0.8ex}{PanEcho}
        & \shortstack{5.189 (4.931--5.460)\\{\scriptsize [$p<0.001$]}}
        & \shortstack{48.443 (42.989--54.830)\\{\scriptsize [$p<0.001$]}}
        & \shortstack{4.873 (4.538--5.219)\\{\scriptsize [$p=0.038$]}}
        & \shortstack{50.381 (40.918--62.726)\\{\scriptsize [$p=0.170$]}}
        & \shortstack{7.404 (7.154--7.661)\\{\scriptsize [$p<0.001$]}}
        & \shortstack{91.690 (85.663--98.400)\\{\scriptsize [$p<0.001$]}} \\
    \midrule
    \raisebox{0.8ex}{DINOv3}
        & \shortstack{5.503 (5.210--5.799)\\{\scriptsize [$p<0.001$]}}
        & \shortstack{55.670 (49.425--62.317)\\{\scriptsize [$p<0.001$]}}
        & \shortstack{5.417 (5.078--5.799)\\{\scriptsize [$p<0.001$]}}
        & \shortstack{61.603 (51.558--73.512)\\{\scriptsize [$p<0.001$]}}
        & \shortstack{7.711 (7.437--7.981)\\{\scriptsize [$p<0.001$]}}
        & \shortstack{99.624 (92.316--106.838)\\{\scriptsize [$p<0.001$]}} \\
    \midrule
    \raisebox{0.8ex}{EchoDino (Uniform)}
        & \shortstack{5.042 (4.776--5.302)\\{\scriptsize [$p<0.001$]}}
        & \shortstack{47.045 (41.935--52.437)\\{\scriptsize [$p<0.001$]}}
        & \shortstack{4.800 (4.474--5.151)\\{\scriptsize [$p=0.006$]}}
        & \shortstack{49.933 (40.817--60.543)\\{\scriptsize [$p=0.036$]}}
        & \shortstack{6.410 (6.173--6.657)\\{\scriptsize [$p<0.001$]}}
        & \shortstack{70.234 (64.792--76.656)\\{\scriptsize [$p<0.001$]}} \\
    \addlinespace[1.5pt]
    EchoDino-MEMS
        & \bf 4.674 (4.456--4.894)
        & \bf 39.353 (35.239--43.259)
        & \bf 4.555 (4.235--4.877)
        & 45.197 (36.002--55.891)
        & \bf 5.652 (5.444--5.854)
        & \bf 56.148 (51.326--61.083) \\
    \bottomrule
    \end{tabular}%
    }
\end{table}

\begin{figure}[!htbp]
    \centering
    \includegraphics[width=0.99\textwidth]{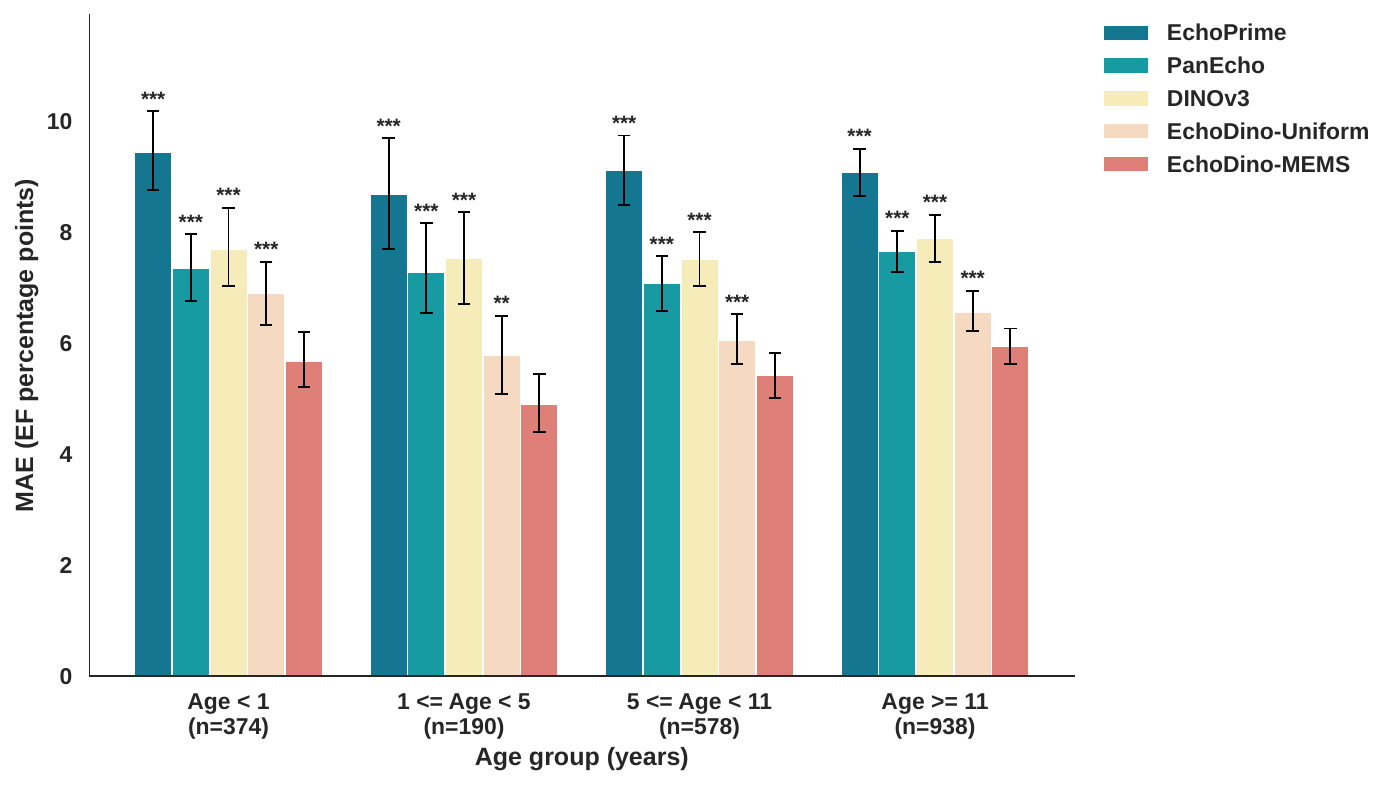}
    \caption{\textbf{Age-stratified ejection-fraction prediction on the TCH-EF test set.} Bootstrap median mean absolute error (MAE, EF percentage points; lower is better) is shown for videos from participants aged $<1$ year ($n=374$), $1$--$<5$ years ($n=190$), $5$--$<11$ years ($n=578$), and $\geq11$ years ($n=938$). EchoDino using uniform sampling achieved lower MAE than the comparison methods in every age group, and EchoDino-MEMS further reduced MAE across all groups. Error bars indicate percentile 95\% confidence intervals. Asterisks denote paired comparisons with EchoDino-MEMS ($\ast\ast: P<0.01$, $\ast\ast\ast: P<0.001$).}
    \label{fig:EF_AgeGroup}
\end{figure}

\textbf{EchoDino and MEMS retain ejection-fraction prediction advantages across pediatric development.} To determine whether the cohort-level advantages persisted throughout pediatric development, we stratified the TCH-EF test set into four age intervals. With uniform sampling, EchoDino achieved lower MAE than all comparison methods in every age group. The comparison between EchoDino and DINOv3 isolates the benefit of adaptation to pediatric echocardiography. EchoDino using uniform sampling reduced MAE across all age groups compared to the baseline models (Fig.~\ref{fig:EF_AgeGroup}). These consistent gains show that adaptation to echocardiography makes EF-relevant changes in ventricular geometry and function more accessible to the shallow temporal readout despite age-dependent variation in cardiac appearance.

MEMS then added a distinct benefit by using the same EchoDino features to favor high-motion, feature-diverse frames while temporal anchors retained whole-video coverage. Relative to EchoDino using uniform sampling, MEMS reduced MAE by 17.9\%, 15.1\%, 10.3\%, and 9.5\% in participants aged $<1$, 1--$<5$, 5--$<11$, and $\geq11$ years, respectively (all $P\leq0.01$). The largest improvement occurred in infants younger than one year ($P<0.001$), a setting in which high heart rates compress phase-defining intervals. Thus, the subgroup analysis demonstrates that EchoDino provides a stable feature substrate across developmental ages, whereas MEMS translates that substrate into improved capture of brief, informative cardiac states.

\textbf{EchoDino and MEMS capture probe-driven acquisition dynamics.} We next tested whether the representation and motion-aware sampling strategy extended beyond cardiac function to sweep-mode classification, which distinguishes four acquisition types (two-dimensional, color Doppler, two-dimensional sweep, and color-Doppler sweep) based on how the ultrasound probe traverses the heart. Unlike EF prediction, this task depends on probe-driven motion distributed across the acquisition rather than on brief phase-specific cardiac states. On TCH-Sweep, EchoDino using uniform sampling achieved an accuracy of 0.957 and a weighted F1 score of 0.955 (Table~\ref{tab:mode_tchSweep_sample}), outperforming PanEcho (0.944 accuracy) and DINOv3 (0.938 accuracy). Because each example contains a sustained, continuous sweep, uniformly spaced frames already capture substantial acquisition-wide information. Nevertheless, MEMS further increased accuracy to 0.961 and weighted F1 score to 0.960. The AUC increased from 0.983 to 0.991 ($p<0.001$), and EchoDino-MEMS significantly outperformed PanEcho across all five evaluated metrics ($p\leq0.05$). Together, these results show that EchoDino features retain information about probe-driven acquisition dynamics, while MEMS adds value by preferentially selecting informative changes rather than only brief cardiac phases.

\begin{table}[htbp]
    \caption{\textbf{Sweep-mode classification on the TCH-Sweep dataset.} Values are bootstrap medians (95\% CIs) of support-weighted metrics. P values compare each method with EchoDino-MEMS.}
    \label{tab:mode_tchSweep_sample}
    \centering
    \small
    \setlength{\tabcolsep}{3pt}
    \resizebox{\textwidth}{!}{%
    \begin{tabular}{lccccc}
    \toprule
    \bf Method & \bf Accuracy $\uparrow$ & \bf Precision $\uparrow$ & \bf Recall $\uparrow$ & \bf F1 score $\uparrow$ & \bf AUC $\uparrow$ \\
    \midrule
    \raisebox{0.8ex}{EchoPrime}
        & \shortstack{0.674 (0.653--0.695)\\{\scriptsize [$p<0.001$]}}
        & \shortstack{0.712 (0.691--0.730)\\{\scriptsize [$p<0.001$]}}
        & \shortstack{0.674 (0.653--0.695)\\{\scriptsize [$p<0.001$]}}
        & \shortstack{0.686 (0.666--0.705)\\{\scriptsize [$p<0.001$]}}
        & \shortstack{0.839 (0.826--0.852)\\{\scriptsize [$p<0.001$]}} \\
    \addlinespace[1.5pt]
    \raisebox{0.8ex}{PanEcho}
        & \shortstack{0.944 (0.934--0.954)\\{\scriptsize [$p<0.001$]}}
        & \shortstack{0.942 (0.931--0.952)\\{\scriptsize [$p<0.001$]}}
        & \shortstack{0.944 (0.934--0.954)\\{\scriptsize [$p<0.001$]}}
        & \shortstack{0.940 (0.928--0.951)\\{\scriptsize [$p<0.001$]}}
        & \shortstack{0.987 (0.983--0.990)\\{\scriptsize [$p=0.010$]}} \\
    \midrule
    \raisebox{0.8ex}{DINOv3}
        & \shortstack{0.938 (0.927--0.948)\\{\scriptsize [$p<0.001$]}}
        & \shortstack{0.939 (0.927--0.949)\\{\scriptsize [$p<0.001$]}}
        & \shortstack{0.938 (0.927--0.948)\\{\scriptsize [$p<0.001$]}}
        & \shortstack{0.938 (0.927--0.949)\\{\scriptsize [$p<0.001$]}}
        & \shortstack{\bf 0.991 (0.988--0.993)\\{\scriptsize [$p=0.914$]}} \\
    \midrule
    \raisebox{0.8ex}{EchoDino (Uniform)}
        & \shortstack{0.957 (0.948--0.965)\\{\scriptsize [$p=0.316$]}}
        & \shortstack{0.955 (0.946--0.964)\\{\scriptsize [$p=0.196$]}}
        & \shortstack{0.957 (0.948--0.965)\\{\scriptsize [$p=0.316$]}}
        & \shortstack{0.955 (0.946--0.963)\\{\scriptsize [$p=0.174$]}}
        & \shortstack{0.983 (0.978--0.988)\\{\scriptsize [$p<0.001$]}} \\
    \addlinespace[1.5pt]
    EchoDino-MEMS
        & \bf 0.961 (0.953--0.969)
        & \bf 0.960 (0.951--0.969)
        & \bf 0.961 (0.953--0.969)
        & \bf 0.960 (0.951--0.969)
        & \bf 0.991 (0.987--0.994) \\
    \bottomrule
    \end{tabular}%
    }
\end{table}

\clearpage
\section{Discussion}
The central finding of this study is that a single, strictly frozen self-supervised representation adapted to pediatric echocardiography can seamlessly support a broad spectrum of visual, anatomical, and functional tasks. Conventionally, automated echocardiographic analysis has relied on fragmented, task-specific models or massive annotated datasets that constrain feature expressivity and scale poorly across heterogeneous clinical workflows. Task-specific models manage this by narrowing the feature space, which in turn requires a large and relatively homogeneous labeled cohort for each task. Cohorts of that size and uniformity are uncommon in pediatric cardiology, and most published models have consequently remained at proof of concept. EchoDino shifts the burden to the representation. By continually adapting DINOv3 via self-supervised pretraining on 3.7 million unlabeled pediatric frames, the network reorganizes its feature space around recurring cardiac geometry and kinetic patterns without relying on downstream diagnostic labels. Consequently, across nine pediatric and adult datasets, this shared representation supported global view recognition, localized measurement, dense segmentation, developmental age estimation, study-level structural heart disease (SHD) classification, and ejection fraction (EF) prediction through lightweight task-specific readouts. These results establish that a unified visual foundation can retain the fine-grained spatial and temporal evidence required for multi-task echocardiographic analysis, eliminating the need to retrain or redesign the core image encoder for each downstream application.

The frame-level experiments clarify how downstream performance depends on matching readout architectures to the spatial organization of the frozen tokens. High view-classification accuracy via $k$-nearest-neighbor and linear probing confirms that global acquisition context is readily accessible from the class token. However, relying solely on global readouts compresses critical spatial detail. By introducing EchoDino-PATCH, we demonstrated that spatially resolved patch tokens natively preserve fine-grained local geometry. Accessing these patch tokens significantly reduced errors in cardiac measurement and age estimation, while intermediate feature hierarchies enabled precise chamber and myocardial boundary segmentation. Crucially, this confirms that EchoDino's frozen tokens function as an expressive visual substrate: local anatomical detail is intrinsically preserved, but harvesting it requires spatial routing tailored to the geometric demands of the clinical target.

Our video-level evaluation demonstrates that dynamic temporal modeling can be achieved without making the foundational encoder intrinsically video-dependent. Independently encoded frame features provided a robust foundation for temporal reasoning, supporting EF regression via a shallow temporal module and study-level SHD diagnosis through prediction aggregation. Furthermore, by introducing MEMS, we addressed a fundamental flaw in conventional video pipelines: content-agnostic uniform frame sampling, which risks omitting brief cardiac phases while expending computational budget on visually redundant frames. MEMS corrects this by prioritizing temporally diverse, high-motion frames, consistently reducing EF prediction errors across evaluated cohorts. This motion-aware selection proved particularly advantageous in infants, where rapid heart rates make capturing brief end-systolic and end-diastolic extremes challenging. Conversely, for sweep-mode acquisitions where discriminative evidence is distributed uniformly across probe traversals, standard sampling remained effective, though MEMS still yielded incremental gains in classification area under the curve (AUC). Together, these findings validate content-aware frame selection as a general mechanism for efficient temporal reasoning in cardiac imaging.

Extending model evaluation to external pediatric and adult benchmarks underscores the broad generalizability of the learned features. Despite pretraining exclusively on pediatric frames, EchoDino's frozen representation transferred effectively to adult datasets, improving ventricular measurement on EchoNet-LVH, tissue segmentation on CAMUS, and EF estimation on EchoNet-Dynamic. This cross-domain success indicates that self-supervised pretraining captures fundamental principles of cardiac anatomy and physiology---such as recurring chamber geometry, tissue boundaries, and phase-dependent contraction mechanics---rather than memorizing cohort-specific demographic identifiers or institutional imaging artifacts. While these findings highlight the potential for tools built on a shared encoder to follow patients transitioning from pediatric to adult congenital heart disease (ACHD) care, direct validation in ACHD populations remains necessary to establish longitudinal measurement consistency as anatomy evolves with age and surgical intervention.

In pediatric cardiology, where marked anatomical heterogeneity, developmental growth, and scarce labeled datasets have long impeded artificial intelligence translation, a reusable foundation model offers a compelling framework. By absorbing developmental and structural variation during label-free pretraining, EchoDino provides a scalable substrate for developing modular clinical tools---ranging from automated acquisition sorting and study-completeness checks to preliminary chamber sizing and functional tracking---without the burden of retraining separate encoders. Crucially, however, the primary value of such automated tools lies in supporting, rather than replacing, expert clinical judgment. Diagnostic evaluation in congenital heart disease is intrinsically integrative, requiring specialists to synthesize spatial relationships, ventricular morphology, loading conditions, surgical history, and complex hemodynamics. Automated extraction of anatomical and functional targets can nonetheless provide a reliable quantitative baseline, reducing repetitive manual measurements, improving inter-examination standardization, and enhancing efficiency during longitudinal surveillance.

Several limitations outline key priorities for future research. First, the primary TCH-SHD endpoint was evaluated as a binary composite classifier combining diverse congenital lesions, valvar regurgitations, and stenoses. Although phenotype-stratified recall analyses indicate broad coverage across underlying conditions, binary detection is fundamentally distinct from detailed anatomical characterization. The current model was not evaluated on sequential segmental analysis, lesion-specific variant discrimination, associated anomalies, or residual postoperative defects. Developing readouts capable of predicting these intermediate anatomical details represents a necessary step toward clinical utility and model explainability.

Second, aggregate performance metrics---such as mean absolute error, Dice coefficients, AUC, and classification accuracy---summarize overall population trends but do not guarantee individual point-of-care actionability. Clinical implementation demands rigorous characterization of measurement bias, limits of agreement, threshold-specific performance, and outlier error rates. This is especially vital in high-risk clinical subgroups, including patients with severe ventricular dysfunction, systemic right or single-ventricle physiology, complex postoperative anatomy, or poor acoustic windows. Finally, for automated tools to safely support clinical decision-making, future iterations must incorporate robust uncertainty estimation and automated quality-control mechanisms capable of flagging degraded or ambiguous examinations where model outputs may be unreliable.

\section{Methods}
To decouple the foundation encoder from individual downstream tasks and assess its transferability across applications, we adapted the self-supervised DINOv3 framework to echocardiography and developed EchoDino (Fig.~\ref{fig:EchoDino_arch}). Self-supervised pretraining does not optimize the encoder for a single clinical label and can produce general-purpose global and localized image representations~\citep{DinoV3}. 

EchoDino is a frame-based foundation model with a ViT~\citep{Vit} backbone. Its frozen encoder provides three complementary representations. The normalized class token summarizes global image context, whereas the grid of normalized final-layer patch tokens retains spatially localized information. Intermediate features extracted across multiple encoder depths preserve fine boundary detail and higher-level anatomy for dense prediction. Together, these representations support static anatomical tasks through lightweight readouts and cardiac-dynamics tasks through shallow temporal fusion, while the backbone remains frozen. In contrast, applying a video foundation model to still-image tasks commonly requires repeating a static frame to construct a pseudo-video~\citep{v-jepa}, an artificial input construction that EchoDino avoids.

\subsection{Domain adaptation of a DINOv3 foundation encoder}
DINOv3 is an image-only self-supervised framework for pretraining vision transformers without text captions or task labels~\citep{DinoV3}. It uses a teacher--student architecture in which the teacher’s predictions supervise the student. The teacher parameters are updated as an exponential moving average of the student parameters, and gradient back-propagation is limited to the student. DINOv3 concurrently learns from whole-image views and masked image patches. Across multiple crops of the same image, it aligns image-level representations to capture global context despite changes in framing, scale, and visible content. The student also predicts teacher features for masked patches, retaining localized and spatially resolved information. During extended training, a dense-feature preservation signal maintains pairwise relationships among patch tokens, supporting spatially coherent representations.

We selected DINOv3 because echocardiographic analysis requires both global and localized representations. View recognition depends on the overall cardiac field of view and acquisition context, whereas segmentation and measurement require spatially coherent features for chambers, ventricular walls, valves, and other anatomical structures. Its multi-crop image-level objective is therefore suited to variation in probe framing and magnification, while masked patch prediction and Gram anchoring support dense anatomical representation learning. EchoDino was initialized from the released DINOv3 ViT-L/16 checkpoint pretrained on LVD-1689M~\cite{DinoV3} and subsequently adapted to 3,747,848 unlabeled pediatric echocardiography frames sampled from 1,681,739 videos.

Although we retained the DINOv3 teacher--student framework, we adapted data construction and augmentation to echocardiographic acquisitions. The student received two $256\times256$ global crops covering 50--100\% of the source frame and eight $96\times96$ local crops covering 10--50\%. The relatively large global crops retained the anatomical context often required for medical-image interpretation, including the spatial relationships among chambers, valves, and the acquisition field of view. The local crops exposed individual cardiac regions at different magnifications, allowing the model to learn localized anatomy while remaining robust to variation in probe framing and zoom. The teacher received the global crops only. Horizontal reflection was disabled because it can reverse cardiac laterality and alter the orientation of standard echocardiographic views. During Gram-anchored continuation, the Gram teacher additionally received an undistorted $512\times512$ view obtained by bicubic resizing, providing a high-resolution spatial reference for preserving dense anatomical relationships. The resulting EchoDino encoder was frozen for all downstream experiments, enabling evaluation of representation transferability without end-to-end task-specific backbone tuning.

\subsection{Frame-level downstream evaluation}
\label{sec: frame-level method}
Frame-level echocardiography tasks require representations at different spatial scales. View recognition depends primarily on the complete acquisition context, whereas cardiac measurement and segmentation require localized anatomical information. We therefore evaluated three readouts from the frozen EchoDino encoder: the final class token, reaggregated final-block patch tokens, and intermediate multiscale patch features.

\textbf{The class-token readout measures task-relevant information directly accessible from EchoDino's global representation.} Each frame was represented by its normalized final class token. For parametric classification and regression, a task-specific linear head mapped this token to the prediction target. View recognition was additionally evaluated using k-nearest neighbors without fitting a parametric head. This class-token readout therefore evaluates global representation quality without task-specific spatial reaggregation or backbone optimization.

\textbf{EchoDino-PATCH reaggregates localized cardiac anatomy when a single global token is insufficient.} The readout concatenated the normalized class token with 256 normalized final-block patch tokens. A four-block, eight-head self-attention decoder allowed the class token to aggregate evidence across the complete $16\times16$ patch grid. Two-dimensional rotary positional encoding preserved patch rows and columns, allowing attention to distinguish anatomically different locations. Final layer normalization and a linear regression head mapped the updated class token to the measurement target. Only the decoder and regression head were optimized. EchoDino-PATCH is therefore a lightweight nonlinear readout, and its comparison with the class-token readout isolates the value of explicit patch access.

\textbf{Intermediate feature hierarchies preserve both boundary detail and high-level anatomy for dense segmentation.} A final-block token grid may omit fine information retained at earlier network depths. Following the DINOv3 linear segmentation protocol, we extracted normalized features from zero-indexed EchoDino transformer blocks $[4,11,17,23]$. The four feature maps were spatially aligned and concatenated along the channel dimension. The segmentation head applied spatial dropout, synchronized batch normalization, and a $1\times1$ convolution to produce patch-grid logits. Bilinear interpolation restored the logits to the input resolution, while the encoder remained frozen. For the EchoPrime baseline, following V-JEPA~\citep{v-jepa}, a single frame was repeated across 16 temporal positions required by its MViTv2~\citep{mVit} backbone. We temporally averaged features from the final transformer blocks of its four resolution stages, at zero-indexed block indices $[0,2,13,15]$. For the PanEcho ConvNeXt~\citep{convnext} backbone, we extracted the four feature-stage outputs at indices $[1,3,5,7]$. These choices yield one output from each resolution stage defined by the hierarchical MViTv2 and ConvNeXt architectures.

\subsection{Video-based temporal fusion}
Video-level echocardiography tasks require spatial evidence from individual frames and temporal evidence across cardiac motion. EchoDino separates these components so that static tasks do not require a video backbone. The frozen frame encoder provides spatial representations, while a lightweight temporal module learns task-specific video aggregation.

\textbf{EchoDino extends frozen frame representations to video prediction through a shallow temporal readout.} Each video was represented by $K=16$ sampled frames. EchoDino independently encoded each frame as one 1,024-dimensional class token and 256 patch tokens. The temporal module averaged the frame-level class tokens to initialize one global sequence token. It then discarded the individual class tokens and flattened all patch tokens across frames. A four-block, eight-head transformer updated the global token by aggregating localized evidence across anatomy and time. Only the temporal module and final regression or classification head were optimized.

\textbf{Multimodal Rotary Positional Embedding (M-RoPE) preserves cardiac location and irregular temporal spacing after frame selection.} Flattening the patch tokens removes their spatial and temporal coordinates. We therefore applied M-RoPE~\cite{mrope} to the attention queries and keys using separate rotary dimensions for patch row, patch column, and source-frame position. For sampled source indices $s_1,\ldots,s_K$, the normalized temporal coordinate was
\begin{equation}
\tau_k=(K-1)\frac{s_k-\min_j s_j}{\max_j s_j-\min_j s_j+\epsilon}.
\end{equation}
We set $\epsilon=10^{-6}$ to avoid division by zero. This transformation maps each video to a common range while retaining relative gaps between nonuniformly selected frames. After four fusion blocks and final layer normalization, the global token was passed to the task head.

\subsection{Motion-biased Entropy Maximization Sampling (MEMS)}
\label{sec:mems}
Uniform sampling provides temporal coverage but may repeatedly select similar cardiac phases or prolonged static intervals. Diversity sampling reduces visual redundancy but can concentrate selections within a temporally narrow portion of a video. MEMS combines temporal anchors, feature-space coverage, and a motion bias to balance these limitations (Fig.~\ref{fig:EchoDino_arch}c,g).

\textbf{Temporal anchors guarantee coarse coverage across the complete echocardiography video.} Given $V=\{v_i\}_{i=0}^{N-1}$ with $N\geq K\geq C$, MEMS first selects $C$ uniformly distributed anchors and samples $K$ frames. The index $c_j$ of anchor $j$ is
\begin{equation}
c_j = j\left\lfloor\frac{N}{C}\right\rfloor + \left\lfloor\frac{N}{2C}\right\rfloor,
\end{equation}
and the seed set is
\begin{equation}
S_{\mathrm{seed}} = \{ v_{c_j} \mid j = 0, 1, \dots, C-1 \}.
\end{equation}
If $C=K$, MEMS reduces to uniform sampling.

\textbf{Frozen EchoDino features quantify frame diversity and cardiac motion without requiring downstream labels.} For each frame $v_i$, we compute the L2-normalized class-token embedding
\begin{equation}
x_i = \frac{\phi(v_i)}{\|\phi(v_i)\|_2}.
\end{equation}
Feature-space motion is the distance between adjacent embeddings. The first observed change is assigned to both initial positions so that the score vector contains $N$ elements:
\begin{equation}
\widetilde{m}_i = \|x_i - x_{i-1}\|_2 \quad (i=1,\ldots,N-1),
\qquad \widetilde{m}_0=\widetilde{m}_1.
\end{equation}
We then normalize the motion scores,
\begin{equation}
m_i = \frac{\widetilde{m}_i}{\mu_m+\epsilon},
\qquad
\mu_m = \frac{1}{N}\sum_{j=0}^{N-1}\widetilde{m}_j,
\end{equation}
where $\epsilon=10^{-6}$ provides numerical stability. Mean normalization makes the motion weight less sensitive to the overall magnitude of feature changes. Frame entropy is measured by cosine distance:
\begin{equation}
d(v_a, v_b) = 1 - (x_a \cdot x_b).
\end{equation}

\textbf{Motion-biased farthest-point sampling favors distinct cardiac appearances near larger motion transitions.} Starting from $S=S_{\mathrm{seed}}$, MEMS iteratively selects
\begin{equation}
v_i^* = \arg\max_{v_i \in V \setminus S}
\left[
\left(\min_{v_k \in S} d(v_i, v_k)\right)
\times
(1 + \lambda m_i)
\right],
\end{equation}
where $\lambda$ controls the motion bias. The minimum-distance term greedily expands feature-space coverage and serves as an entropy-maximization surrogate. Multiplication by $(1+\lambda m_i)$ prioritizes frames near larger transitions without removing the diversity constraint. After each selection, the new frame is added to $S$, and all minimum distances are updated. Selection stops when $|S|=K$. The selected indices are sorted chronologically before temporal fusion to preserve acquisition order. Experimentally, we used $K=16$, $C=4$, and $\lambda=0.6$, producing four temporal anchors and twelve motion-biased diversity selections. The ablation study is provided in Supplementary Note~\ref{si-note:mems-ablation} and Supplementary Fig.~\ref{si-fig:EchoDynamic_ablation}.

\section{Experimental setup}
\subsection{Datasets}\label{sec:datasets}
The evaluation was designed to test representation quality across pediatric and adult populations, internal and external institutions, and tasks requiring global, localized, dense, or temporal information. Six datasets were derived from Texas Children's Hospital (TCH; Table~\ref{tab:tch_cohort_summary}). EchoNet-Pediatric provided an external pediatric benchmark, while CAMUS, EchoNet-LVH, and EchoNet-Dynamic provided adult benchmarks.

\textbf{TCH-Complex} provided the unlabeled pretraining corpus. The source archive contained 1,681,739 video records corresponding to 1,680,332 unique DICOM files from 17,984 studies and 13,323 pediatric patients. It included two-dimensional, three-dimensional, color Doppler, M-mode, B-mode, and related acquisitions. We sampled at most three frames from each video to limit repeated presentation of highly similar adjacent frames and prevent longer acquisitions from contributing disproportionately to pretraining. This procedure produced 3,747,848 pretraining images. 

\textbf{TCH downstream cohorts} were curated separately for their respective tasks, as shown in Table~\ref{tab:tch_cohort_summary}. \textbf{TCH-SHD} comprised 1,015,795 retained videos from 16,427 studies in 12,517 patients. At the patient level, 4,736 patients had cardiologist-provided labels for a broad class of significant structural heart disease (select congenital and acquired heart disease), and 7,781 were labeled normal. Twenty-four non-mutually-exclusive diagnostic and phenotypic labels were reserved for post hoc stratification. Detailed label definitions are provided in Supplementary Methods.
\textbf{TCH-View} contained 10,522 retained videos from 301 studies in 298 patients. The videos were assigned to 52 cardiologist-annotated classes that jointly encoded acquisition mode, standard view family, and a reference designation, anatomical level, or focus.
\textbf{TCH-EF} contained 13,809 apical four-chamber videos from 3,864 studies in 2,637 patients, paired with study-level EF. Mean EF was \(46.59\pm14.60\%\), with a range of 0.27--79.00\%.
\textbf{TCH-Age} retained the first frame from each eligible video and comprised 101,636 frames from 3,865 studies in 2,915 patients aged 0--18 years. The frames represented the 11 two-dimensional view categories reported in Supplementary Table~\ref{si-tab:tch_age_composition}. The parasternal categories were the short-axis view (\texttt{2D\_PSAX}) and long-axis view (\texttt{2D\_PLAX}). The apical categories were the four-chamber (\texttt{2D\_A4C}), two-chamber (\texttt{2D\_A2C}), five-chamber (\texttt{2D\_A5C}), and three-chamber (\texttt{2D\_A3C}) views. The subcostal categories were the long-axis view (\texttt{2D\_SCLAX}) and short-axis views acquired with the indicator at 12~o'clock (\texttt{2D\_SCSAX12}) or 6~o'clock (\texttt{2D\_SCSAX6}). The suprasternal-notch categories were the long-axis view (\texttt{2D\_SSNLAX}) and short-axis view (\texttt{2D\_SSNSAX}). For TCH-Age only, frame-level predictions were aggregated within each patient before evaluation.
\textbf{TCH-Sweep} contained 13,410 videos from 316 studies in 313 patients and distinguished four acquisition modes. The cohort included 5,582 two-dimensional, 6,209 color-Doppler, 723 two-dimensional-sweep, and 896 color-Doppler-sweep videos. Detailed class and target distributions are provided in Supplementary Methods and Supplementary Tables~\ref{si-tab:tch_view_distribution}--\ref{si-tab:tch_sweep_composition}.
TCH-SHD is a subset of the TCH-Complex dataset, consisting of 56.8\% DICOM files and 57.4\% studies. The remaining TCH downstream cohorts contained no studies included in TCH-Complex.

\textbf{EchoNet-Pediatric} \citep{EchoPed} contains 7,643 videos from 4,467 examinations in 1,958 patients younger than 18 years. The dataset provides apical four-chamber and parasternal short-axis videos with LV endocardial tracings at end diastole and end systole and clinical EF calculated using the pediatric 5/6 area--length method. We used it for frame-level LV segmentation and video-level EF prediction.

\textbf{CAMUS} \citep{CAMUS} contains apical two- and four-chamber sequences from 500 adults examined at the University Hospital of Saint-Etienne. Cardiologist annotations identify end-diastolic and end-systolic frames and delineate the LV endocardium, LV epicardium, and left atrium. We used CAMUS for frame-level segmentation of the LV cavity, myocardium, and left atrium.

\textbf{EchoNet-LVH} \citep{echoNetLVH} contains 12,000 adult parasternal long-axis videos with an official split of 9,600/1,200/1,200 videos for training, validation, and testing. Caliper annotations identify the interventricular septum, LV internal dimension, and LV posterior wall. We evaluated LVID at end diastole (LVIDd) and end systole (LVIDs), posterior-wall thickness at end diastole (LVPWd), and septal thickness at end diastole (IVSd).

\textbf{EchoNet-Dynamic} \citep{EchoDyanmic} contains 10,030 labeled adult apical four-chamber videos with LV tracings and clinical measurements of EF, end-systolic volume, and end-diastolic volume. We used it for frame-level volume regression, frame-level LV segmentation, and video-level EF prediction.

We used official partitions for public datasets if available. For TCH cohorts without predefined partitions, we constructed patient-disjoint, label-stratified training, validation, and test sets; cohort-specific partition sizes are reported in Supplementary Tables~\ref{si-tab:tch_view_distribution}--\ref{si-tab:tch_sweep_composition}.

\subsection{Preprocessing, optimization, and compute}
\textbf{Image overlays and metadata exclusion.} For TCH-Complex and all downstream TCH cohorts, videos were provided to the model only as pixel data after masking the top image overlay bar. No DICOM header fields, tabular metadata, patient identifiers, timestamps, scanner or vendor information, acquisition descriptions, or free-text reports were supplied as model inputs. These data were used only where necessary to assemble cohorts, define task labels, link videos within studies and patients, and construct patient-disjoint splits; they were not used as predictive features, input embeddings, or frame-selection signals.

\textbf{EchoDino Pretraining} Each frame was converted to a three-channel image and normalized using the ImageNet mean $(0.485,0.456,0.406)$ and standard deviation $(0.229,0.224,0.225)$. The student received two $256\times256$ global crops covering 50--100\% of the source frame and eight $96\times96$ local crops covering 10--50\%. Horizontal reflection was disabled to preserve cardiac laterality and view orientation. During Gram-anchored continuation, the Gram teacher additionally received an undistorted $512\times512$ view obtained by bicubic resizing. The DINOv3 ViT-L/16 weights initialized both student and teacher networks. We used AdamW with $\beta_1=0.9$, $\beta_2=0.999$, a base learning rate of $10^{-3}$, a 10,000-step warm-up, cosine decay to $10^{-6}$, and weight decay increasing from 0.04 to 0.4. Training used 48 images per GPU across four NVIDIA A40 GPUs, yielding 192 images per training step. Gradients were accumulated for five steps before each optimizer update, producing an effective optimizer batch size of 960 images. Model parameters were stored in bfloat16, distributed reductions were performed in float32, and gradients were clipped to a norm of 3.0. The Gram-anchored continuation used a learning rate increasing to $3\times10^{-5}$, weight decay of 0.04, teacher momentum of 0.999, and a gradient-norm threshold of 30. EchoDino is trained for 150,000 steps for the first stage and another 150,000 steps for the second stage with the Gram Loss. The complete pretraining run required approximately 672 GPU-hours.

\textbf{Downstream Tasks} Frames were resized without changing their aspect ratio and zero-padded to $256\times256$ for EchoDino or $224\times224$ for PanEcho and EchoPrime. Training images were randomly rotated by up to $\pm15^\circ$; validation and test images received no stochastic augmentation. EchoDino and PanEcho inputs were scaled to $[0,1]$ and normalized using ImageNet statistics. EchoPrime inputs retained the $[0,255]$ intensity scale and used its prescribed channel mean and standard deviation. For frame-level EchoPrime experiments, each image was repeated across the 16 temporal positions required by its video backbone. We optimized only the task-specific readout for 30 epochs using AdamW, weight decay of $10^{-3}$, a three-epoch linear warm-up followed by cosine learning-rate decay, bfloat16 mixed precision, and gradient clipping at a norm of 1.0. Learning rates were selected on the validation set from task-specific grids spanning $5\times10^{-5}$ to $10^{-3}$, with the same candidates applied to every encoder compared on a given task. Training used two NVIDIA A6000 GPU processes and a fixed random seed of 42. Per-device batches were typically 128 for frame-level EchoDino and PanEcho experiments and 32 for video-level experiments. EchoPrime used a batch size of 16 because its 16-frame input requires substantially more memory, including when a still frame is repeated for frame-level evaluation.

\textbf{Baseline models.} We compared EchoDino with DINOv3, PanEcho, and EchoPrime to represent natural-image self-supervision, supervised echocardiography pretraining, and video-native vision--language pretraining, respectively. DINOv3 provided the initialization-matched control for measuring the effect of self-supervised adaptation to pediatric echocardiography \citep{DinoV3}. PanEcho provided an echocardiography-specific frame encoder trained within a view-agnostic multitask system on more than one million videos \citep{PanEcho}. EchoPrime provided the principal video-native comparison through an mViT encoder trained on more than 12 million video--report pairs \citep{EchoPrime}.

\textbf{Evaluation Metrics} Classification metrics included accuracy, precision, recall, F1 score, and area under the receiver operating characteristic curve (AUC), where applicable. Regression metrics included mean absolute error (MAE) and mean squared error (MSE), and segmentation performance was measured using the Dice coefficient. We used 10,000 paired nonparametric bootstrap resamples to estimate median performance, percentile 95\% confidence intervals, and two-sided $p$ values from paired performance differences. Resampling was performed at the independent evaluation unit for each endpoint, with repeated observations grouped before resampling. Identical resamples were applied to every model evaluated on the same test set.

\clearpage
\section*{Ethics}
The dataset from Texas Children’s Hospital/Baylor College of Medicine was collected with institutional review board approval (H-52659) and waiver of informed consent for retrospective analysis of existing deidentified clinical data.

\section*{Data availability}
The TCH-related datasets used in this study are not publicly available because they originate from protected clinical data and remain subject to institutional data-governance requirements. Deidentified data may be made available from the corresponding authors on reasonable request, subject to institutional review, approval, and any applicable data-use agreements.

The EchoNet-Dynamic dataset is available at https://echonet.github.io/dynamic/. 

The EchoNet-Pediatric dataset is available at https://aimi.stanford.edu/datasets/echonet-pediatric. 

The EchoNet-LVH dataset is available at https://echonet.github.io/lvh/. 

The CAMUS dataset is available at https://www.kaggle.com/datasets/shoybhasan/camus-human-heart-data.

\section*{Code availability}
The EchoDino pretraining code and example notebooks are available at
\url{https://github.com/Sean-C71/EchoDino}.

\section*{Acknowledgements}
This work was supported in part by the RANGE cluster operated by Rice University's Center for Research Computing (CRC). M.B.N. is supported by the National Center for Advancing Translational Sciences award K12TR004522.

\section*{Author contributions}
Conceptualization: S.C., D.S., and M.B.N.; Data curation: M.B.N., C.G.R., and D.M.O.; Formal analysis: S.C., D.S., and M.B.N.; Funding acquisition: M.B.N. and D.J.P.; Methodology: S.C., D.S., M.B.N., and D.M.O.; Project administration: D.S. and M.B.N.; Supervision: D.S. and M.B.N.; Validation: D.S., M.B.N., and D.M.O.; Visualization: S.C.; Writing---original draft: S.C.; Writing---review and editing: S.C., D.S., M.B.N., C.G.R., D.J.P., and D.M.O. All authors reviewed the manuscript, contributed important intellectual content, and approved the final version.

\section*{Competing interests}
The authors declare no competing interests.

\clearpage

\bibliographystyle{unsrtnat}
\bibliography{echoDino}

\clearpage
\setcounter{figure}{0}
\setcounter{table}{0}
\setcounter{suppnote}{0}
\renewcommand{\figurename}{Supplementary Fig.}
\renewcommand{\tablename}{Supplementary Table}
\renewcommand{\thefigure}{\arabic{figure}}
\renewcommand{\thetable}{\arabic{table}}
\renewcommand{\theHfigure}{supplementary.\arabic{figure}}
\renewcommand{\theHtable}{supplementary.\arabic{table}}
\makeatletter
\setlength{\@fptop}{0pt}
\setlength{\@fpsep}{14pt}
\setlength{\@fpbot}{0pt plus 1fil}
\makeatother
\renewcommand{\topfraction}{0.95}
\renewcommand{\bottomfraction}{0.95}
\renewcommand{\textfraction}{0.06}
\renewcommand{\floatpagefraction}{0.80}
\setlength{\textfloatsep}{16pt plus 3pt minus 3pt}
\setlength{\floatsep}{14pt plus 3pt minus 3pt}

\begin{center}
{\LARGE\bfseries EchoDino: A pediatric foundation model for transferable echocardiographic analysis across the lifespan\par}
\vspace{1.2em}
{\Large Supplementary Information\par}
\vspace{1.2em}
Sheng~Cheng\textsuperscript{1},
Donnchadh M.~O'Sullivan\textsuperscript{2,3},
Daniel~J.~Penny\textsuperscript{2,3},
Craig~G. Rusin\textsuperscript{2,3},
Minh~B.~Nguyen\textsuperscript{2,3,*}, and
Devika~Subramanian\textsuperscript{1,*,\dag }\\[0.8em]
{\small
\textsuperscript{1}Department of Computer Science, Rice University, Houston, TX, USA\\
\textsuperscript{2}Division of Pediatric Cardiology, Baylor College of Medicine, Houston, TX, USA\\
\textsuperscript{3}Texas Children's Hospital, Houston, TX, USA\\
*Co-senior authors
\textsuperscript{\dag}Correspondence: devika@rice.edu}
\end{center}

\vspace{1.5em}

\section*{Supplementary Methods}
\label{si-sec:supplementary-methods}

\subsection*{TCH-SHD label definition}
\label{si-method:tch-shd-definition}
The structural heart disease phenotype was defined using a prespecified, rule-based combination of structured congenital lesion variables, structured valvar regurgitation grades, and valve-stenosis severity extracted from the echocardiography report summary. The resulting SHD labels were reviewed by an expert cardiologist. An examination was classified as positive for structural heart disease if it demonstrated at least one congenital structural cardiac lesion, moderate or severe pulmonary, tricuspid, mitral, or aortic regurgitation, or moderate-or-greater pulmonary, aortic, mitral, or tricuspid stenosis.

Congenital structural heart disease was defined by the presence of at least one lesion-level diagnosis, including d-transposition of the great arteries, congenitally corrected transposition of the great arteries, coarctation of the aorta, interrupted aortic arch, truncus arteriosus, aortopulmonary window, total anomalous pulmonary venous connection, tetralogy of Fallot, double-outlet right ventricle, double-outlet left ventricle, complete or other atrioventricular septal defect, tricuspid atresia, pulmonary atresia, pulmonary atresia with intact ventricular septum, critical pulmonary stenosis, critical aortic stenosis, Ebstein anomaly, hypoplastic left heart syndrome, hypoplastic left ventricle, hypoplastic right ventricle, double-inlet left ventricle, double-inlet right ventricle, atrial septal defect, ventricular septal defect, or patent ductus arteriosus. Moderate or severe valvar regurgitation was identified from structured regurgitation grades. Moderate-or-greater valvar stenosis was identified using valve-specific extraction of stenosis severity from the echocardiography report summary. Examinations meeting none of the congenital lesion, clinically significant valvar regurgitation, or moderate-or-greater valvar stenosis criteria were classified as negative. Postoperative status alone did not determine the phenotype; repaired or palliated congenital heart disease remained classified as positive when the underlying congenital diagnosis was present.

\subsection*{Task-specific label and target distributions}

The following tables summarize the label and target distributions used in the five downstream TCH cohorts. Supplementary Table~\ref{si-tab:tch_view_distribution} reports cohort sizes, partition-specific frequencies, standardized clinical names, and display codes for the 52 retained TCH-View acquisition classes. Supplementary Tables~\ref{si-tab:tch_chd_distribution}--\ref{si-tab:tch_sweep_composition} report the task-specific composition of TCH-SHD, TCH-EF, TCH-Age, and TCH-Sweep. Counts and summary statistics are reported at the video, study, or patient level specified in each table.

\begingroup
\scriptsize
\setlength{\tabcolsep}{1.2pt}
\renewcommand{\arraystretch}{1.05}
\begin{longtable}{@{}>{\raggedright\arraybackslash}p{4.45cm} >{\raggedright\arraybackslash}p{2.75cm} *{4}{>{\raggedleft\arraybackslash}p{1.55cm}}@{}}
\caption{\textbf{Composition, video-level class distribution and nomenclature crosswalk for the TCH-View dataset.} Cohort sizes report retained videos, unique studies and unique patients within each partition. Class percentages are calculated within each partition. The 52 retained display codes encode acquisition mode, view family and a reference designation, anatomical level or focus; designated reference acquisitions are identified by the suffix REF. Every retained class occurred in all three patient- and study-disjoint partitions.}
\label{si-tab:tch_view_distribution}\\
\toprule
\textbf{Standardized clinical name} & \textbf{Display code} & \textbf{Training} & \textbf{Validation} & \textbf{Test} & \textbf{Total} \\
\midrule
\endfirsthead
\multicolumn{6}{l}{\tablename~\thetable\ continued} \\
\toprule
\textbf{Standardized clinical name} & \textbf{Display code} & \textbf{Training} & \textbf{Validation} & \textbf{Test} & \textbf{Total} \\
\midrule
\endhead
\midrule
\multicolumn{6}{r}{Continued on the next page} \\
\endfoot
\bottomrule
\endlastfoot
\multicolumn{2}{@{}l}{\textit{Cohort size}} & & & & \\
\addlinespace[1pt]
\multicolumn{2}{@{}l}{Retained videos, $n$} & 8,577 & 967 & 978 & 10,522 \\
\multicolumn{2}{@{}l}{Studies, $n$} & 244 & 29 & 28 & 301 \\
\multicolumn{2}{@{}l}{Patients, $n$} & 242 & 28 & 28 & 298 \\
\midrule
\multicolumn{6}{@{}l}{\textit{Video-level class distribution, $n$ (\%)}} \\
\addlinespace[1pt]
Two-dimensional parasternal short-axis reference view at the aortic-valve level & \nolinkurl{2D-PSAX-AV-REF} & \mbox{678 (7.9\%)} & \mbox{62 (6.4\%)} & \mbox{86 (8.8\%)} & \mbox{826 (7.9\%)} \\
Two-dimensional apical four-chamber reference view & \nolinkurl{2D-A4C-REF} & \mbox{612 (7.1\%)} & \mbox{82 (8.5\%)} & \mbox{80 (8.2\%)} & \mbox{774 (7.4\%)} \\
Color-Doppler parasternal short-axis aortic-valve-level view, coronary-artery focus & \nolinkurl{CD-PSAX-AV-COR} & \mbox{508 (5.9\%)} & \mbox{52 (5.4\%)} & \mbox{22 (2.2\%)} & \mbox{582 (5.5\%)} \\
Two-dimensional parasternal long-axis reference view & \nolinkurl{2D-PLAX-REF} & \mbox{457 (5.3\%)} & \mbox{55 (5.7\%)} & \mbox{51 (5.2\%)} & \mbox{563 (5.4\%)} \\
Two-dimensional parasternal short-axis view at the mitral-valve/papillary-muscle level & \nolinkurl{2D-PSAX-MVPM} & \mbox{440 (5.1\%)} & \mbox{58 (6.0\%)} & \mbox{61 (6.2\%)} & \mbox{559 (5.3\%)} \\
Color-Doppler parasternal long-axis reference view & \nolinkurl{CD-PLAX-REF} & \mbox{412 (4.8\%)} & \mbox{41 (4.2\%)} & \mbox{48 (4.9\%)} & \mbox{501 (4.8\%)} \\
Color-Doppler parasternal short-axis view focused on the right-ventricular outflow tract, pulmonary valve, and main pulmonary artery & \nolinkurl{CD-PSAX-RVOT} & \mbox{364 (4.2\%)} & \mbox{39 (4.0\%)} & \mbox{38 (3.9\%)} & \mbox{441 (4.2\%)} \\
Color-Doppler apical five-chamber left-ventricular-outflow-tract reference view & \nolinkurl{CD-A5C-LVOT-REF} & \mbox{270 (3.1\%)} & \mbox{28 (2.9\%)} & \mbox{32 (3.3\%)} & \mbox{330 (3.1\%)} \\
Two-dimensional apical two-chamber reference view & \nolinkurl{2D-A2C-REF} & \mbox{255 (3.0\%)} & \mbox{37 (3.8\%)} & \mbox{24 (2.5\%)} & \mbox{316 (3.0\%)} \\
Color-Doppler left-heart-focused apical four-chamber view & \nolinkurl{CD-A4C-LH} & \mbox{243 (2.8\%)} & \mbox{25 (2.6\%)} & \mbox{29 (3.0\%)} & \mbox{297 (2.8\%)} \\
Two-dimensional parasternal short-axis view focused on the right-ventricular outflow tract, pulmonary valve, and main pulmonary artery & \nolinkurl{2D-PSAX-RVOT} & \mbox{230 (2.7\%)} & \mbox{25 (2.6\%)} & \mbox{26 (2.7\%)} & \mbox{281 (2.7\%)} \\
Color-Doppler suprasternal long-axis aortic-arch view & \nolinkurl{CD-SS-LAX-ARCH} & \mbox{232 (2.7\%)} & \mbox{25 (2.6\%)} & \mbox{23 (2.4\%)} & \mbox{280 (2.7\%)} \\
Color-Doppler apical right-ventricle-centered view & \nolinkurl{CD-AP-RVC} & \mbox{219 (2.6\%)} & \mbox{20 (2.1\%)} & \mbox{25 (2.6\%)} & \mbox{264 (2.5\%)} \\
Color-Doppler parasternal long-axis view focused on the right-ventricular outflow tract, pulmonary valve, and main pulmonary artery & \nolinkurl{CD-PLAX-RVOT} & \mbox{203 (2.4\%)} & \mbox{20 (2.1\%)} & \mbox{25 (2.6\%)} & \mbox{248 (2.4\%)} \\
Color-Doppler parasternal short-axis reference view at the aortic-valve level & \nolinkurl{CD-PSAX-AV-REF} & \mbox{202 (2.4\%)} & \mbox{20 (2.1\%)} & \mbox{23 (2.4\%)} & \mbox{245 (2.3\%)} \\
Color-Doppler parasternal long-axis tricuspid-valve/right-ventricular-inflow view & \nolinkurl{CD-PLAX-TV} & \mbox{194 (2.3\%)} & \mbox{21 (2.2\%)} & \mbox{23 (2.4\%)} & \mbox{238 (2.3\%)} \\
Two-dimensional suprasternal long-axis aortic-arch view & \nolinkurl{2D-SS-LAX-ARCH} & \mbox{192 (2.2\%)} & \mbox{16 (1.7\%)} & \mbox{22 (2.2\%)} & \mbox{230 (2.2\%)} \\
Two-dimensional apical five-chamber left-ventricular-outflow-tract reference view & \nolinkurl{2D-A5C-LVOT-REF} & \mbox{176 (2.1\%)} & \mbox{17 (1.8\%)} & \mbox{25 (2.6\%)} & \mbox{218 (2.1\%)} \\
Color-Doppler subcostal sagittal inferior-vena-cava/descending-aorta view, descending-aorta focus & \nolinkurl{CD-SC-SAG-DAo} & \mbox{170 (2.0\%)} & \mbox{19 (2.0\%)} & \mbox{21 (2.1\%)} & \mbox{210 (2.0\%)} \\
Two-dimensional parasternal long-axis tricuspid-valve/right-ventricular-inflow view & \nolinkurl{2D-PLAX-TV} & \mbox{165 (1.9\%)} & \mbox{18 (1.9\%)} & \mbox{25 (2.6\%)} & \mbox{208 (2.0\%)} \\
Color-Doppler subcostal sagittal inferior-vena-cava/descending-aorta view, inferior-vena-cava focus & \nolinkurl{CD-SC-SAG-IVC} & \mbox{176 (2.1\%)} & \mbox{15 (1.6\%)} & \mbox{16 (1.6\%)} & \mbox{207 (2.0\%)} \\
Color-Doppler parasternal short-axis view, tricuspid-valve focus & \nolinkurl{CD-PSAX-TV} & \mbox{157 (1.8\%)} & \mbox{16 (1.7\%)} & \mbox{23 (2.4\%)} & \mbox{196 (1.9\%)} \\
Color-Doppler parasternal short-axis view, pulmonary-vein focus & \nolinkurl{CD-PSAX-PVein} & \mbox{177 (2.1\%)} & \mbox{13 (1.3\%)} & \mbox{3 (0.3\%)} & \mbox{193 (1.8\%)} \\
Two-dimensional parasternal long-axis view focused on the right-ventricular outflow tract, pulmonary valve, and main pulmonary artery & \nolinkurl{2D-PLAX-RVOT} & \mbox{153 (1.8\%)} & \mbox{20 (2.1\%)} & \mbox{14 (1.4\%)} & \mbox{187 (1.8\%)} \\
Color-Doppler subcostal coronal (long-axis) reference view & \nolinkurl{CD-SC-COR-LAX-REF} & \mbox{142 (1.7\%)} & \mbox{15 (1.6\%)} & \mbox{11 (1.1\%)} & \mbox{168 (1.6\%)} \\
Color-Doppler suprasternal short-axis pulmonary-vein ('crab') view & \nolinkurl{CD-SS-SAX-PVein} & \mbox{98 (1.1\%)} & \mbox{30 (3.1\%)} & \mbox{17 (1.7\%)} & \mbox{145 (1.4\%)} \\
Color-Doppler subcostal sagittal (short-axis) reference view & \nolinkurl{CD-SC-SAG-SAX-REF} & \mbox{119 (1.4\%)} & \mbox{12 (1.2\%)} & \mbox{13 (1.3\%)} & \mbox{144 (1.4\%)} \\
Color-Doppler apical four-chamber view, atrial-septum focus & \nolinkurl{CD-A4C-AS} & \mbox{108 (1.3\%)} & \mbox{18 (1.9\%)} & \mbox{11 (1.1\%)} & \mbox{137 (1.3\%)} \\
Two-dimensional apical three-chamber reference view & \nolinkurl{2D-A3C-REF} & \mbox{106 (1.2\%)} & \mbox{17 (1.8\%)} & \mbox{11 (1.1\%)} & \mbox{134 (1.3\%)} \\
Two-dimensional subcostal coronal (long-axis) reference view & \nolinkurl{2D-SC-COR-LAX-REF} & \mbox{113 (1.3\%)} & \mbox{8 (0.8\%)} & \mbox{12 (1.2\%)} & \mbox{133 (1.3\%)} \\
Two-dimensional subcostal sagittal inferior-vena-cava/descending-aorta view, inferior-vena-cava focus & \nolinkurl{2D-SC-SAG-IVC} & \mbox{90 (1.0\%)} & \mbox{7 (0.7\%)} & \mbox{10 (1.0\%)} & \mbox{107 (1.0\%)} \\
Two-dimensional left-heart-focused apical four-chamber view & \nolinkurl{2D-A4C-LH} & \mbox{70 (0.8\%)} & \mbox{9 (0.9\%)} & \mbox{18 (1.8\%)} & \mbox{97 (0.9\%)} \\
Two-dimensional apical right-ventricle-centered view & \nolinkurl{2D-AP-RVC} & \mbox{73 (0.9\%)} & \mbox{8 (0.8\%)} & \mbox{13 (1.3\%)} & \mbox{94 (0.9\%)} \\
Color-Doppler suprasternal long-axis view, ascending-aorta focus & \nolinkurl{CD-SS-LAX-AAo} & \mbox{76 (0.9\%)} & \mbox{12 (1.2\%)} & \mbox{6 (0.6\%)} & \mbox{94 (0.9\%)} \\
Color-Doppler parasternal short-axis view at the mitral-valve/papillary-muscle level & \nolinkurl{CD-PSAX-MVPM} & \mbox{70 (0.8\%)} & \mbox{12 (1.2\%)} & \mbox{9 (0.9\%)} & \mbox{91 (0.9\%)} \\
Two-dimensional parasternal short-axis view, tricuspid-valve focus & \nolinkurl{2D-PSAX-TV} & \mbox{69 (0.8\%)} & \mbox{5 (0.5\%)} & \mbox{10 (1.0\%)} & \mbox{84 (0.8\%)} \\
Two-dimensional parasternal short-axis view at the apical level & \nolinkurl{2D-PSAX-APEX} & \mbox{58 (0.7\%)} & \mbox{12 (1.2\%)} & \mbox{9 (0.9\%)} & \mbox{79 (0.8\%)} \\
Two-dimensional subcostal sagittal inferior-vena-cava/descending-aorta view, descending-aorta focus & \nolinkurl{2D-SC-SAG-DAo} & \mbox{63 (0.7\%)} & \mbox{5 (0.5\%)} & \mbox{9 (0.9\%)} & \mbox{77 (0.7\%)} \\
Color-Doppler apical two-chamber reference view & \nolinkurl{CD-A2C-REF} & \mbox{49 (0.6\%)} & \mbox{7 (0.7\%)} & \mbox{10 (1.0\%)} & \mbox{66 (0.6\%)} \\
Color-Doppler apical three-chamber reference view & \nolinkurl{CD-A3C-REF} & \mbox{58 (0.7\%)} & \mbox{5 (0.5\%)} & \mbox{3 (0.3\%)} & \mbox{66 (0.6\%)} \\
Color-Doppler suprasternal short-axis view, pulmonary-artery focus & \nolinkurl{CD-SS-SAX-PA} & \mbox{38 (0.4\%)} & \mbox{6 (0.6\%)} & \mbox{10 (1.0\%)} & \mbox{54 (0.5\%)} \\
Two-dimensional suprasternal short-axis view, aortic-arch/branch-vessel focus & \nolinkurl{2D-SS-SAX-ARCH} & \mbox{43 (0.5\%)} & \mbox{2 (0.2\%)} & \mbox{4 (0.4\%)} & \mbox{49 (0.5\%)} \\
Two-dimensional subcostal sagittal (short-axis) reference view & \nolinkurl{2D-SC-SAG-SAX-REF} & \mbox{39 (0.5\%)} & \mbox{3 (0.3\%)} & \mbox{6 (0.6\%)} & \mbox{48 (0.5\%)} \\
Color-Doppler anteriorly angled apical five-chamber view focused on the right-ventricular outflow tract and pulmonary valve & \nolinkurl{CD-A5C-RVOT-PVal} & \mbox{37 (0.4\%)} & \mbox{8 (0.8\%)} & \mbox{3 (0.3\%)} & \mbox{48 (0.5\%)} \\
Color-Doppler subcostal sagittal (short-axis) view, atrial-septum focus & \nolinkurl{CD-SC-SAG-SAX-AS} & \mbox{32 (0.4\%)} & \mbox{6 (0.6\%)} & \mbox{1 (0.1\%)} & \mbox{39 (0.4\%)} \\
Color-Doppler parasternal short-axis view, branch-pulmonary-artery focus & \nolinkurl{CD-PSAX-BPA} & \mbox{28 (0.3\%)} & \mbox{1 (0.1\%)} & \mbox{7 (0.7\%)} & \mbox{36 (0.3\%)} \\
Color-Doppler suprasternal short-axis view, aortic-arch/branch-vessel focus & \nolinkurl{CD-SS-SAX-ARCH} & \mbox{27 (0.3\%)} & \mbox{1 (0.1\%)} & \mbox{1 (0.1\%)} & \mbox{29 (0.3\%)} \\
Color-Doppler subcostal coronal (long-axis) view, aortic focus & \nolinkurl{CD-SC-COR-LAX-Ao} & \mbox{24 (0.3\%)} & \mbox{3 (0.3\%)} & \mbox{1 (0.1\%)} & \mbox{28 (0.3\%)} \\
Two-dimensional suprasternal short-axis view, pulmonary-artery focus & \nolinkurl{2D-SS-SAX-PA} & \mbox{14 (0.2\%)} & \mbox{6 (0.6\%)} & \mbox{1 (0.1\%)} & \mbox{21 (0.2\%)} \\
Color-Doppler subcostal sagittal (short-axis) view, superior-vena-cava focus & \nolinkurl{CD-SC-SAG-SAX-SVC} & \mbox{17 (0.2\%)} & \mbox{1 (0.1\%)} & \mbox{3 (0.3\%)} & \mbox{21 (0.2\%)} \\
Two-dimensional subcostal coronal situs view & \nolinkurl{2D-SC-COR-SITUS} & \mbox{16 (0.2\%)} & \mbox{3 (0.3\%)} & \mbox{1 (0.1\%)} & \mbox{20 (0.2\%)} \\
Two-dimensional subcostal sagittal (short-axis) view at the mitral-valve/papillary-muscle level & \nolinkurl{2D-SC-SAG-SAX-MVPM} & \mbox{15 (0.2\%)} & \mbox{1 (0.1\%)} & \mbox{3 (0.3\%)} & \mbox{19 (0.2\%)} \\
\end{longtable}
\endgroup

\begin{table}[htbp]
\centering
\caption{\textbf{Composition and binary label distribution of TCH-SHD.} Cohort sizes report retained videos, unique studies, and unique patients within each partition. Label counts are reported separately at the video, study, and patient levels, with percentages calculated within each partition and data level. Patients were mutually exclusive across partitions.}
\label{si-tab:tch_chd_distribution}
\footnotesize
\setlength{\tabcolsep}{3pt}
\renewcommand{\arraystretch}{1.08}
\begin{tabularx}{\textwidth}{@{}>{\raggedright\arraybackslash}p{3.2cm} *{4}{>{\centering\arraybackslash}X}@{}}
\toprule
\textbf{Characteristic} & \textbf{Training} & \textbf{Validation} & \textbf{Test} & \textbf{Total} \\
\midrule
\multicolumn{5}{@{}l}{\textit{Cohort size}} \\
\addlinespace[1pt]
Retained videos, $n$ & 810,362 & 103,008 & 102,425 & 1,015,795 \\
Studies, $n$ & 13,114 & 1,676 & 1,637 & 16,427 \\
Patients, $n$ & 10,013 & 1,252 & 1,252 & 12,517 \\
\midrule
\multicolumn{5}{@{}l}{\textit{Video-level label distribution, $n$ (\%)}} \\
\addlinespace[1pt]
Normal & 484,465 (59.8\%) & 62,478 (60.7\%) & 62,709 (61.2\%) & 609,652 (60.0\%) \\
SHD\_any & 325,897 (40.2\%) & 40,530 (39.3\%) & 39,716 (38.8\%) & 406,143 (40.0\%) \\
\midrule
\multicolumn{5}{@{}l}{\textit{Study-level label distribution, $n$ (\%)}} \\
\addlinespace[1pt]
Normal & 7,425 (56.6\%) & 958 (57.2\%) & 941 (57.5\%) & 9,324 (56.8\%) \\
SHD\_any & 5,689 (43.4\%) & 718 (42.8\%) & 696 (42.5\%) & 7,103 (43.2\%) \\
\midrule
\multicolumn{5}{@{}l}{\textit{Patient-level label distribution, $n$ (\%)}} \\
\addlinespace[1pt]
Normal & 6,224 (62.2\%) & 778 (62.1\%) & 779 (62.2\%) & 7,781 (62.2\%) \\
SHD\_any & 3,789 (37.8\%) & 474 (37.9\%) & 473 (37.8\%) & 4,736 (37.8\%) \\
\bottomrule
\end{tabularx}
\end{table}

\begin{table}[htbp]
\centering
\caption{\textbf{Composition and label distribution of TCH-EF.} Cohort sizes report retained videos, unique studies, and unique patients within each partition. Age and ejection fraction (EF) summaries are calculated at the study level. Partitions were generated at the patient level using ten equal-frequency bins of patient-level median EF; no patient or study occurred in more than one partition. IQR denotes interquartile range and SD denotes standard deviation.}
\label{si-tab:tch_ef_composition}
\footnotesize
\setlength{\tabcolsep}{3pt}
\renewcommand{\arraystretch}{1.08}
\begin{tabularx}{\textwidth}{@{}>{\raggedright\arraybackslash}p{3.2cm} *{4}{>{\centering\arraybackslash}X}@{}}
\toprule
\textbf{Characteristic} & \textbf{Training} & \textbf{Validation} & \textbf{Test} & \textbf{Total} \\
\midrule
\multicolumn{5}{@{}l}{\textit{Cohort size}} \\
\addlinespace[1pt]
Retained videos, $n$ & 9,441 & 2,288 & 2,080 & 13,809 \\
Studies, $n$ & 2,614 & 666 & 584 & 3,864 \\
Patients, $n$ & 1,792 & 449 & 396 & 2,637 \\
\midrule
\multicolumn{5}{@{}l}{\textit{Video-level view distribution, $n$ (\%)}} \\
\addlinespace[1pt]
\texttt{A4C\_ref\_2D} & 9,441 (100.0\%) & 2,288 (100.0\%) & 2,080 (100.0\%) & 13,809 (100.0\%) \\
\midrule
\multicolumn{5}{@{}l}{\textit{Age distribution}} \\
\addlinespace[1pt]
Mean $\pm$ SD, years & $9.30\pm6.25$ & $8.77\pm6.28$ & $9.16\pm6.25$ & $9.19\pm6.26$ \\
Median (IQR), years & 10.09 (3.09--15.04) & 9.36 (1.36--14.56) & 9.97 (2.24--14.77) & 9.92 (2.68--14.92) \\
Range, years & 0.00--18.99 & 0.00--18.88 & 0.00--18.99 & 0.00--18.99 \\
\midrule
\multicolumn{5}{@{}l}{\textit{EF distribution}} \\
\addlinespace[1pt]
Mean $\pm$ SD, \% & $46.56\pm14.69$ & $46.46\pm14.66$ & $46.87\pm14.12$ & $46.59\pm14.60$ \\
Median (IQR), \% & 47.39 (37.00--58.27) & 47.09 (36.06--57.58) & 47.00 (37.07--58.35) & 47.24 (36.99--58.10) \\
Range, \% & 0.27--78.00 & 7.00--79.00 & 0.44--74.00 & 0.27--79.00 \\
\bottomrule
\end{tabularx}
\end{table}

\begin{table}[htbp]
\centering
\caption{\textbf{Composition of TCH-Age.} Cohort sizes report retained frames, unique studies, and unique patients within each partition. View distributions are frame-level counts, with percentages calculated within each partition. Age summaries are calculated at the study level. Studies and patients were mutually exclusive across partitions. IQR denotes interquartile range and SD denotes standard deviation.}
\label{si-tab:tch_age_composition}
\footnotesize
\setlength{\tabcolsep}{3pt}
\renewcommand{\arraystretch}{1.08}
\begin{tabularx}{\textwidth}{@{}>{\raggedright\arraybackslash}p{3.2cm} *{4}{>{\centering\arraybackslash}X}@{}}
\toprule
\textbf{Characteristic} & \textbf{Training} & \textbf{Validation} & \textbf{Test} & \textbf{Total} \\
\midrule
\multicolumn{5}{@{}l}{\textit{Cohort size}} \\
\addlinespace[1pt]
Retained frames, $n$ & 81,619 & 10,059 & 9,958 & 101,636 \\
Studies, $n$ & 3,095 & 385 & 385 & 3,865 \\
Patients, $n$ & 2,196 & 349 & 370 & 2,915 \\
\midrule
\multicolumn{5}{@{}l}{\textit{Frame-level view distribution, $n$ (\%)}} \\
\addlinespace[1pt]
\texttt{2D\_PSAX} & 26,011 (31.9\%) & 3,255 (32.4\%) & 3,212 (32.3\%) & 32,478 (32.0\%) \\
\texttt{2D\_A4C} & 16,445 (20.1\%) & 1,983 (19.7\%) & 2,014 (20.2\%) & 20,442 (20.1\%) \\
\texttt{2D\_PLAX} & 13,149 (16.1\%) & 1,600 (15.9\%) & 1,607 (16.1\%) & 16,356 (16.1\%) \\
\texttt{2D\_A2C} & 5,768 (7.1\%) & 729 (7.2\%) & 701 (7.0\%) & 7,198 (7.1\%) \\
\texttt{2D\_SCLAX} & 5,203 (6.4\%) & 637 (6.3\%) & 639 (6.4\%) & 6,479 (6.4\%) \\
\texttt{2D\_SCSAX12} & 2,906 (3.6\%) & 387 (3.8\%) & 345 (3.5\%) & 3,638 (3.6\%) \\
\texttt{2D\_A5C} & 2,776 (3.4\%) & 345 (3.4\%) & 363 (3.6\%) & 3,484 (3.4\%) \\
\texttt{2D\_A3C} & 2,750 (3.4\%) & 326 (3.2\%) & 325 (3.3\%) & 3,401 (3.3\%) \\
\texttt{2D\_SSNLAX} & 2,636 (3.2\%) & 318 (3.2\%) & 301 (3.0\%) & 3,255 (3.2\%) \\
\texttt{2D\_SCSAX6} & 2,440 (3.0\%) & 306 (3.0\%) & 273 (2.7\%) & 3,019 (3.0\%) \\
\texttt{2D\_SSNSAX} & 1,535 (1.9\%) & 173 (1.7\%) & 178 (1.8\%) & 1,886 (1.9\%) \\
\midrule
\multicolumn{5}{@{}l}{\textit{Age distribution}} \\
\addlinespace[1pt]
Mean $\pm$ SD, years & $8.75\pm6.07$ & $8.73\pm6.07$ & $8.85\pm6.10$ & $8.76\pm6.07$ \\
Median (IQR), years & 9.39 (2.17--14.37) & 9.31 (2.08--14.47) & 9.55 (2.09--14.57) & 9.44 (2.15--14.42) \\
Range, years & 0.00--17.99 & 0.00--18.00 & 0.00--17.99 & 0.00--18.00 \\
\bottomrule
\end{tabularx}
\end{table}

\begin{table}[htbp]
\centering
\caption{\textbf{Composition of TCH-Sweep by data partition.} Partitions were constructed at the patient level with stratification by acquisition mode; patients and studies were mutually exclusive across partitions. Acquisition-mode distributions are video-level counts, with percentages calculated within each partition.}
\label{si-tab:tch_sweep_composition}
\footnotesize
\setlength{\tabcolsep}{3pt}
\renewcommand{\arraystretch}{1.08}
\begin{tabularx}{\textwidth}{@{}>{\raggedright\arraybackslash}p{4.2cm} *{4}{>{\centering\arraybackslash}X}@{}}
\toprule
\textbf{Characteristic} & \textbf{Training} & \textbf{Validation} & \textbf{Test} & \textbf{Total} \\
\midrule
\multicolumn{5}{@{}l}{\textit{Cohort size}} \\
\addlinespace[1pt]
Videos, $n$ & 9,388 & 2,011 & 2,011 & 13,410 \\
Studies, $n$ & 219 & 50 & 47 & 316 \\
Patients, $n$ & 219 & 47 & 47 & 313 \\
\midrule
\multicolumn{5}{@{}l}{\textit{Acquisition-mode distribution, $n$ (\%)}} \\
\addlinespace[1pt]
Color Doppler & 4,345 (46.3\%) & 932 (46.3\%) & 932 (46.3\%) & 6,209 (46.3\%) \\
Two-dimensional & 3,908 (41.6\%) & 837 (41.6\%) & 837 (41.6\%) & 5,582 (41.6\%) \\
Color-Doppler sweep & 628 (6.7\%) & 134 (6.7\%) & 134 (6.7\%) & 896 (6.7\%) \\
Two-dimensional sweep & 507 (5.4\%) & 108 (5.4\%) & 108 (5.4\%) & 723 (5.4\%) \\
\bottomrule
\end{tabularx}
\end{table}

\clearpage
\subsection*{Scanner-vendor coverage}

Scanner-vendor metadata were retained to describe acquisition heterogeneity. Scanner-model labels were considerably more granular and sparsely represented, so model-level rows were omitted. Counts are non-mutually-exclusive when a study or patient contains examinations from more than one system.

\begin{table}[htbp]
\centering
\caption{\textbf{Scanner-vendor coverage across the TCH cohorts.} Vendor counts denote unique studies associated with each metadata label. Counts need not sum to the cohort total because vendor metadata can be missing and one study can contain acquisitions from more than one system. The two TomTec spellings in TCH-Complex are preserved as separate source labels.}
\label{si-tab:tch_vendor_distribution}
\small
\renewcommand{\arraystretch}{1.12}
\begin{tabular}{@{}lp{11.6cm}@{}}
\toprule
\textbf{Dataset} & \textbf{Vendor-associated studies} \\
\midrule
TCH-Complex & GE Vingmed Ultrasound, 16,386; Philips Medical Systems, 1,622; TomTec Imaging Systems GmbH, 700; TOMTEC, 126 \\
TCH-SHD & GE Vingmed Ultrasound, 15,007; Philips Medical Systems, 1,437; TomTec Imaging Systems GmbH, 189 \\
TCH-View & GE Vingmed Ultrasound, 186; Philips Medical Systems, 94; TOSHIBA\_MEC\_US, 23; SIEMENS, 7; GEMS Ultrasound, 5; GE Healthcare, 1 \\
TCH-EF & GE Vingmed Ultrasound, 3,053; Philips Medical Systems, 716; TOSHIBA\_MEC\_US, 196; SIEMENS, 33; GEMS Ultrasound, 28; GE Healthcare, 4 \\
TCH-Age & GE Vingmed Ultrasound, 3,066; Philips Medical Systems, 719; TOSHIBA\_MEC\_US, 198; SIEMENS, 34; GEMS Ultrasound, 28; GE Healthcare, 4 \\
TCH-Sweep & GE Vingmed Ultrasound, 186; Philips Medical Systems, 94; TOSHIBA\_MEC\_US, 23; SIEMENS, 6; GEMS Ultrasound, 5; GE Healthcare, 1 \\
\bottomrule
\end{tabular}
\end{table}

\clearpage
\begin{landscape}
\suppnote{Detailed view-classification analysis}{si-note:view-classification}
The TCH-View cohort contained 52 combined acquisition-mode, standard-view-family, and reference/level/focus classes. Supplementary Table~\ref{si-tab:view_classification_per_class} reports bootstrap distributions of precision, recall, and F1 score for the retained classes, followed by support-weighted results. Standardized display codes are used in Supplementary Table~\ref{si-tab:view_classification_per_class}, with designated reference acquisitions explicitly identified by the suffix REF; the corresponding original acquisition labels and full clinical names are provided in Supplementary Table~\ref{si-tab:tch_view_distribution}.

\begingroup
\tiny
\setlength{\tabcolsep}{1.2pt}
\renewcommand{\arraystretch}{1.12}
\begin{longtable}{@{}>{\raggedright\arraybackslash}m{2.8cm} >{\centering\arraybackslash}m{0.45cm} *{12}{>{\centering\arraybackslash}m{1.35cm}}@{}}
\caption{\textbf{Bootstrap analysis of per-class linear-probe view classification on TCH-View.} The display-code column uses the standardized codes defined in Supplementary Table~\ref{si-tab:tch_view_distribution}, which provides the corresponding clinical names. Each cell reports the median, 95\% confidence interval, and bootstrap $p$ value from 10,000 resamples. EchoDino is the reference model for the reported $p$ values, and its $p$-value fields are marked by \textnormal{[-]}. Within each row and metric, the highest median is bolded and the second-highest distinct median is underlined; tied values receive the same formatting. A dash in a precision cell indicates that the model made no positive predictions for that class, such that the precision denominator was zero and the metric was undefined. P, R, and F1 denote precision, recall, and F1 score, respectively, and $n$ denotes test support.}
\label{si-tab:view_classification_per_class}\\
\toprule
& & \multicolumn{3}{c}{\bf EchoPrime} & \multicolumn{3}{c}{\bf PanEcho} & \multicolumn{3}{c}{\bf DINOv3} & \multicolumn{3}{c}{\bf EchoDino} \\
\cmidrule(lr){3-5}\cmidrule(lr){6-8}\cmidrule(lr){9-11}\cmidrule(lr){12-14}
\bf Display code & \bf $n$ & \bf P & \bf R & \bf F1 & \bf P & \bf R & \bf F1 & \bf P & \bf R & \bf F1 & \bf P & \bf R & \bf F1 \\
\midrule
\endfirsthead
\multicolumn{14}{l}{\tablename~\thetable\ continued} \\
\toprule
& & \multicolumn{3}{c}{\bf EchoPrime} & \multicolumn{3}{c}{\bf PanEcho} & \multicolumn{3}{c}{\bf DINOv3} & \multicolumn{3}{c}{\bf EchoDino} \\
\cmidrule(lr){3-5}\cmidrule(lr){6-8}\cmidrule(lr){9-11}\cmidrule(lr){12-14}
\bf Display code & \bf $n$ & \bf P & \bf R & \bf F1 & \bf P & \bf R & \bf F1 & \bf P & \bf R & \bf F1 & \bf P & \bf R & \bf F1 \\
\midrule
\endhead
\midrule
\multicolumn{14}{r}{Continued on the next page} \\
\endfoot
\bottomrule
\endlastfoot
\nolinkurl{2D-A2C-REF} & 24 & 0.143 (0.058--0.246) [$p<0.001$] & 0.333 (0.147--0.533) [$p<0.001$] & 0.200 (0.085--0.323) [$p<0.001$] & \underline{0.731} (0.526--0.905) [$p=0.005$] & \underline{0.667} (0.469--0.852) [$p=0.002$] & \underline{0.696} (0.519--0.833) [$p<0.001$] & 0.306 (0.167--0.459) [$p<0.001$] & 0.500 (0.294--0.706) [$p<0.001$] & 0.379 (0.218--0.526) [$p<0.001$] & \textbf{1.000} (1.000--1.000) [-] & \textbf{0.962} (0.857--1.000) [-] & \textbf{0.980} (0.923--1.000) [-] \\
\nolinkurl{2D-A3C-REF} & 11 & 0.000 (0.000--0.000) [$p<0.001$] & 0.000 (0.000--0.000) [$p<0.001$] & 0.000 (0.000--0.000) [$p<0.001$] & 0.100 (0.000--0.375) [$p<0.001$] & 0.083 (0.000--0.308) [$p<0.001$] & 0.095 (0.000--0.300) [$p<0.001$] & \underline{0.222} (0.000--0.500) [$p<0.001$] & \underline{0.267} (0.000--0.571) [$p<0.001$] & \underline{0.240} (0.000--0.476) [$p<0.001$] & \textbf{1.000} (1.000--1.000) [-] & \textbf{1.000} (1.000--1.000) [-] & \textbf{1.000} (1.000--1.000) [-] \\
\nolinkurl{2D-A4C-LH} & 18 & - & 0.000 (0.000--0.000) [$p<0.001$] & 0.000 (0.000--0.000) [$p<0.001$] & \underline{0.333} (0.000--1.000) [$p=0.258$] & \underline{0.053} (0.000--0.188) [$p<0.001$] & \underline{0.091} (0.000--0.286) [$p<0.001$] & - & 0.000 (0.000--0.000) [$p<0.001$] & 0.000 (0.000--0.000) [$p<0.001$] & \textbf{0.722} (0.455--0.933) [-] & \textbf{0.556} (0.316--0.789) [-] & \textbf{0.625} (0.400--0.800) [-] \\
\nolinkurl{2D-A4C-REF} & 80 & 0.342 (0.272--0.415) [$p<0.001$] & 0.738 (0.638--0.831) [$p=0.048$] & 0.467 (0.388--0.542) [$p<0.001$] & \underline{0.525} (0.434--0.615) [$p<0.001$] & 0.776 (0.679--0.863) [$p=0.149$] & 0.626 (0.542--0.701) [$p<0.001$] & 0.519 (0.432--0.605) [$p<0.001$] & \underline{0.839} (0.753--0.914) [$p=0.712$] & \underline{0.641} (0.560--0.713) [$p<0.001$] & \textbf{0.833} (0.747--0.908) [-] & \textbf{0.864} (0.783--0.934) [-] & \textbf{0.847} (0.783--0.902) [-] \\
\nolinkurl{2D-AP-RVC} & 13 & - & 0.000 (0.000--0.000) [$p<0.001$] & 0.000 (0.000--0.000) [$p<0.001$] & \textbf{0.667} (0.000--1.000) [$p=0.934$] & \underline{0.143} (0.000--0.385) [$p=0.004$] & \underline{0.235} (0.000--0.526) [$p=0.021$] & 0.000 (0.000--0.000) [$p<0.001$] & 0.000 (0.000--0.000) [$p<0.001$] & 0.000 (0.000--0.000) [$p<0.001$] & \underline{0.647} (0.400--0.875) [-] & \textbf{0.857} (0.615--1.000) [-] & \textbf{0.733} (0.519--0.889) [-] \\
\nolinkurl{2D-A5C-LVOT-REF} & 25 & - & 0.000 (0.000--0.000) [$p<0.001$] & 0.000 (0.000--0.000) [$p<0.001$] & 0.286 (0.095--0.500) [$p<0.001$] & \underline{0.238} (0.080--0.421) [$p<0.001$] & \underline{0.256} (0.091--0.426) [$p<0.001$] & \underline{0.500} (0.000--1.000) [$p=0.250$] & 0.074 (0.000--0.200) [$p<0.001$] & 0.129 (0.000--0.320) [$p<0.001$] & \textbf{0.889} (0.750--1.000) [-] & \textbf{0.926} (0.800--1.000) [-] & \textbf{0.905} (0.800--0.979) [-] \\
\nolinkurl{2D-PLAX-RVOT} & 14 & - & 0.000 (0.000--0.000) [$p<0.001$] & 0.000 (0.000--0.000) [$p<0.001$] & \underline{0.545} (0.222--0.857) [$p=0.013$] & \underline{0.429} (0.167--0.706) [$p<0.001$] & \underline{0.476} (0.200--0.700) [$p<0.001$] & - & 0.000 (0.000--0.000) [$p<0.001$] & 0.000 (0.000--0.000) [$p<0.001$] & \textbf{1.000} (1.000--1.000) [-] & \textbf{1.000} (1.000--1.000) [-] & \textbf{1.000} (1.000--1.000) [-] \\
\nolinkurl{2D-PLAX-REF} & 51 & 0.345 (0.222--0.475) [$p<0.001$] & 0.372 (0.241--0.510) [$p<0.001$] & 0.357 (0.237--0.472) [$p<0.001$] & \underline{0.855} (0.750--0.941) [$p=0.030$] & \underline{0.905} (0.812--0.978) [$p=0.013$] & \underline{0.877} (0.803--0.937) [$p=0.001$] & 0.500 (0.373--0.627) [$p<0.001$] & 0.589 (0.450--0.723) [$p<0.001$] & 0.540 (0.421--0.647) [$p<0.001$] & \textbf{0.964} (0.902--1.000) [-] & \textbf{1.000} (1.000--1.000) [-] & \textbf{0.982} (0.949--1.000) [-] \\
\nolinkurl{2D-PLAX-TV} & 25 & - & 0.000 (0.000--0.000) [$p<0.001$] & 0.000 (0.000--0.000) [$p<0.001$] & \underline{0.700} (0.500--0.880) [$p=0.007$] & \underline{0.643} (0.444--0.828) [$p<0.001$] & \underline{0.667} (0.488--0.808) [$p<0.001$] & - & 0.000 (0.000--0.000) [$p<0.001$] & 0.000 (0.000--0.000) [$p<0.001$] & \textbf{0.964} (0.870--1.000) [-] & \textbf{1.000} (1.000--1.000) [-] & \textbf{0.982} (0.930--1.000) [-] \\
\nolinkurl{2D-PSAX-MVPM} & 61 & 0.441 (0.323--0.561) [$p<0.001$] & \underline{0.492} (0.365--0.618) [$p<0.001$] & 0.464 (0.352--0.567) [$p<0.001$] & \underline{0.800} (0.701--0.885) [$p<0.001$] & \textbf{0.969} (0.915--1.000) [$p=1.000$] & \underline{0.875} (0.809--0.929) [$p<0.001$] & 0.440 (0.357--0.525) [$p<0.001$] & \textbf{0.969} (0.915--1.000) [$p=1.000$] & 0.605 (0.519--0.682) [$p<0.001$] & \textbf{0.969} (0.915--1.000) [-] & \textbf{0.969} (0.915--1.000) [-] & \textbf{0.968} (0.930--0.993) [-] \\
\nolinkurl{2D-PSAX-APEX} & 9 & - & 0.000 (0.000--0.000) [$p=0.001$] & 0.000 (0.000--0.000) [$p=0.001$] & \underline{0.667} (0.000--1.000) [$p=0.857$] & \underline{0.214} (0.000--0.545) [$p=0.013$] & \underline{0.333} (0.000--0.667) [$p=0.027$] & - & 0.000 (0.000--0.000) [$p=0.001$] & 0.000 (0.000--0.000) [$p=0.001$] & \textbf{0.800} (0.455--1.000) [-] & \textbf{0.800} (0.444--1.000) [-] & \textbf{0.778} (0.500--0.952) [-] \\
\nolinkurl{2D-PSAX-RVOT} & 26 & 0.000 (0.000--0.000) [$p<0.001$] & 0.000 (0.000--0.000) [$p<0.001$] & 0.000 (0.000--0.000) [$p<0.001$] & 0.435 (0.231--0.647) [$p<0.001$] & \underline{0.382} (0.200--0.579) [$p<0.001$] & \underline{0.407} (0.222--0.571) [$p<0.001$] & \underline{0.600} (0.000--1.000) [$p=0.292$] & 0.111 (0.000--0.250) [$p<0.001$] & 0.188 (0.000--0.385) [$p<0.001$] & \textbf{0.889} (0.750--1.000) [-] & \textbf{0.889} (0.750--1.000) [-] & \textbf{0.889} (0.776--0.964) [-] \\
\nolinkurl{2D-PSAX-AV-REF} & 86 & 0.273 (0.217--0.332) [$p<0.001$] & 0.745 (0.649--0.835) [$p=0.002$] & 0.400 (0.330--0.466) [$p<0.001$] & \underline{0.680} (0.588--0.767) [$p<0.001$] & \underline{0.839} (0.755--0.912) [$p=0.008$] & \underline{0.750} (0.677--0.815) [$p<0.001$] & 0.560 (0.465--0.653) [$p<0.001$] & 0.711 (0.610--0.804) [$p<0.001$] & 0.626 (0.541--0.702) [$p<0.001$] & \textbf{0.943} (0.886--0.988) [-] & \textbf{0.932} (0.871--0.978) [-] & \textbf{0.936} (0.894--0.970) [-] \\
\nolinkurl{2D-PSAX-TV} & 10 & - & 0.000 (0.000--0.000) [$p<0.001$] & 0.000 (0.000--0.000) [$p<0.001$] & \textbf{1.000} (1.000--1.000) [$p=0.736$] & \underline{0.091} (0.000--0.333) [$p=0.002$] & \underline{0.167} (0.000--0.500) [$p=0.003$] & - & 0.000 (0.000--0.000) [$p<0.001$] & 0.000 (0.000--0.000) [$p<0.001$] & \underline{0.900} (0.625--1.000) [-] & \textbf{0.812} (0.500--1.000) [-] & \textbf{0.846} (0.600--1.000) [-] \\
\nolinkurl{2D-SC-SAG-DAo} & 9 & - & 0.000 (0.000--0.000) [$p=0.004$] & 0.000 (0.000--0.000) [$p=0.004$] & \underline{0.833} (0.333--1.000) [$p=0.724$] & \underline{0.444} (0.111--0.800) [$p=0.268$] & \underline{0.571} (0.182--0.857) [$p=0.099$] & - & 0.000 (0.000--0.000) [$p=0.004$] & 0.000 (0.000--0.000) [$p=0.004$] & \textbf{1.000} (1.000--1.000) [-] & \textbf{0.667} (0.333--1.000) [-] & \textbf{0.800} (0.500--1.000) [-] \\
\nolinkurl{2D-SC-SAG-IVC} & 10 & - & 0.000 (0.000--0.000) [$p<0.001$] & 0.000 (0.000--0.000) [$p<0.001$] & \underline{0.667} (0.333--1.000) [$p=0.527$] & \underline{0.600} (0.250--0.900) [$p=0.035$] & \underline{0.632} (0.308--0.857) [$p=0.071$] & 0.600 (0.000--1.000) [$p=0.521$] & 0.286 (0.000--0.625) [$p=0.001$] & 0.400 (0.000--0.667) [$p=0.012$] & \textbf{0.778} (0.500--1.000) [-] & \textbf{1.000} (1.000--1.000) [-] & \textbf{0.875} (0.667--1.000) [-] \\
\nolinkurl{2D-SC-SAG-SAX-MVPM} & 3 & - & \underline{0.000} (0.000--0.000) [$p<0.001$] & \underline{0.000} (0.000--0.000) [$p<0.001$] & - & \underline{0.000} (0.000--0.000) [$p<0.001$] & \underline{0.000} (0.000--0.000) [$p<0.001$] & - & \underline{0.000} (0.000--0.000) [$p<0.001$] & \underline{0.000} (0.000--0.000) [$p<0.001$] & \textbf{1.000} (1.000--1.000) [-] & \textbf{1.000} (1.000--1.000) [-] & \textbf{1.000} (1.000--1.000) [-] \\
\nolinkurl{2D-SC-SAG-SAX-REF} & 6 & - & 0.000 (0.000--0.000) [$p<0.001$] & 0.000 (0.000--0.000) [$p<0.001$] & \underline{0.571} (0.167--1.000) [$p=0.098$] & \underline{0.667} (0.200--1.000) [$p=0.264$] & \underline{0.615} (0.200--0.889) [$p=0.013$] & - & 0.000 (0.000--0.000) [$p<0.001$] & 0.000 (0.000--0.000) [$p<0.001$] & \textbf{1.000} (1.000--1.000) [-] & \textbf{1.000} (1.000--1.000) [-] & \textbf{1.000} (1.000--1.000) [-] \\
\nolinkurl{2D-SS-LAX-ARCH} & 22 & 0.222 (0.057--0.417) [$p<0.001$] & 0.222 (0.059--0.421) [$p<0.001$] & 0.222 (0.057--0.391) [$p<0.001$] & \underline{0.684} (0.474--0.875) [$p=0.002$] & \underline{0.684} (0.474--0.875) [$p=0.002$] & \underline{0.682} (0.500--0.824) [$p<0.001$] & 0.533 (0.267--0.800) [$p=0.002$] & 0.364 (0.167--0.576) [$p<0.001$] & 0.429 (0.211--0.619) [$p<0.001$] & \textbf{1.000} (1.000--1.000) [-] & \textbf{1.000} (1.000--1.000) [-] & \textbf{1.000} (1.000--1.000) [-] \\
\nolinkurl{2D-SS-SAX-ARCH} & 4 & - & \underline{0.000} (0.000--0.000) [$p<0.001$] & \underline{0.000} (0.000--0.000) [$p<0.001$] & - & \underline{0.000} (0.000--0.000) [$p<0.001$] & \underline{0.000} (0.000--0.000) [$p<0.001$] & - & \underline{0.000} (0.000--0.000) [$p<0.001$] & \underline{0.000} (0.000--0.000) [$p<0.001$] & \textbf{1.000} (1.000--1.000) [-] & \textbf{1.000} (1.000--1.000) [-] & \textbf{1.000} (1.000--1.000) [-] \\
\nolinkurl{2D-SS-SAX-PA} & 1 & - & \underline{0.000} (0.000--0.000) [$p<0.001$] & \underline{0.000} (0.000--0.000) [$p<0.001$] & - & \underline{0.000} (0.000--0.000) [$p<0.001$] & \underline{0.000} (0.000--0.000) [$p<0.001$] & - & \underline{0.000} (0.000--0.000) [$p<0.001$] & \underline{0.000} (0.000--0.000) [$p<0.001$] & \textbf{1.000} (1.000--1.000) [-] & \textbf{1.000} (1.000--1.000) [-] & \textbf{1.000} (1.000--1.000) [-] \\
\nolinkurl{2D-SC-COR-SITUS} & 1 & - & \underline{0.000} (0.000--0.000) [$p<0.001$] & \underline{0.000} (0.000--0.000) [$p<0.001$] & \underline{0.000} (0.000--0.000) [$p<0.001$] & \underline{0.000} (0.000--0.000) [$p<0.001$] & \underline{0.000} (0.000--0.000) [$p<0.001$] & - & \underline{0.000} (0.000--0.000) [$p<0.001$] & \underline{0.000} (0.000--0.000) [$p<0.001$] & \textbf{1.000} (1.000--1.000) [-] & \textbf{1.000} (1.000--1.000) [-] & \textbf{1.000} (1.000--1.000) [-] \\
\nolinkurl{2D-SC-COR-LAX-REF} & 12 & - & 0.000 (0.000--0.000) [$p<0.001$] & 0.000 (0.000--0.000) [$p<0.001$] & \underline{0.600} (0.250--0.900) [$p=0.036$] & \underline{0.500} (0.200--0.800) [$p=0.005$] & \underline{0.545} (0.235--0.774) [$p<0.001$] & 0.400 (0.000--1.000) [$p=0.085$] & 0.154 (0.000--0.417) [$p<0.001$] & 0.222 (0.000--0.500) [$p<0.001$] & \textbf{1.000} (1.000--1.000) [-] & \textbf{1.000} (1.000--1.000) [-] & \textbf{1.000} (1.000--1.000) [-] \\
\nolinkurl{CD-A2C-REF} & 10 & - & \underline{0.000} (0.000--0.000) [$p=0.005$] & \underline{0.000} (0.000--0.000) [$p=0.005$] & \underline{0.000} (0.000--0.000) [$p<0.001$] & \underline{0.000} (0.000--0.000) [$p=0.005$] & \underline{0.000} (0.000--0.000) [$p=0.005$] & - & \underline{0.000} (0.000--0.000) [$p=0.005$] & \underline{0.000} (0.000--0.000) [$p=0.005$] & \textbf{1.000} (1.000--1.000) [-] & \textbf{0.600} (0.250--0.900) [-] & \textbf{0.750} (0.400--0.947) [-] \\
\nolinkurl{CD-A3C-REF} & 3 & - & 0.000 (0.000--0.000) [$p<0.001$] & 0.000 (0.000--0.000) [$p<0.001$] & \textbf{1.000} (1.000--1.000) [$p=1.000$] & \underline{0.333} (0.000--1.000) [$p=0.177$] & \underline{0.500} (0.000--1.000) [$p=0.177$] & - & 0.000 (0.000--0.000) [$p<0.001$] & 0.000 (0.000--0.000) [$p<0.001$] & \textbf{1.000} (1.000--1.000) [-] & \textbf{1.000} (1.000--1.000) [-] & \textbf{1.000} (1.000--1.000) [-] \\
\nolinkurl{CD-A4C-AS} & 11 & - & 0.000 (0.000--0.000) [$p=0.099$] & 0.000 (0.000--0.000) [$p=0.099$] & \underline{0.750} (0.000--1.000) [$p=0.714$] & \textbf{0.267} (0.000--0.571) [$p=1.000$] & \underline{0.400} (0.000--0.696) [$p=0.948$] & \textbf{1.000} (1.000--1.000) [$p=1.000$] & \underline{0.083} (0.000--0.300) [$p=0.272$] & 0.154 (0.000--0.462) [$p=0.272$] & \textbf{1.000} (1.000--1.000) [-] & \textbf{0.267} (0.000--0.571) [-] & \textbf{0.421} (0.000--0.727) [-] \\
\nolinkurl{CD-A4C-LH} & 29 & 0.133 (0.000--0.364) [$p<0.001$] & 0.065 (0.000--0.176) [$p<0.001$] & 0.089 (0.000--0.222) [$p<0.001$] & 0.472 (0.308--0.639) [$p<0.001$] & \underline{0.586} (0.400--0.767) [$p=0.001$] & \underline{0.522} (0.361--0.659) [$p<0.001$] & \underline{0.517} (0.333--0.704) [$p=0.005$] & 0.517 (0.333--0.700) [$p<0.001$] & 0.516 (0.345--0.667) [$p<0.001$] & \textbf{0.780} (0.633--0.909) [-] & \textbf{0.968} (0.885--1.000) [-] & \textbf{0.862} (0.759--0.941) [-] \\
\nolinkurl{CD-AP-RVC} & 25 & 0.280 (0.156--0.419) [$p<0.001$] & 0.520 (0.318--0.720) [$p=0.002$] & 0.364 (0.214--0.505) [$p<0.001$] & \underline{0.609} (0.400--0.808) [$p=0.003$] & 0.560 (0.357--0.759) [$p=0.004$] & 0.582 (0.391--0.735) [$p<0.001$] & 0.488 (0.333--0.643) [$p<0.001$] & \underline{0.806} (0.630--0.952) [$p=0.339$] & \underline{0.606} (0.449--0.735) [$p<0.001$] & \textbf{0.889} (0.750--1.000) [-] & \textbf{0.926} (0.795--1.000) [-] & \textbf{0.905} (0.800--0.979) [-] \\
\nolinkurl{CD-A5C-RVOT-PVal} & 3 & - & \underline{0.000} (0.000--0.000) [$p<0.001$] & \underline{0.000} (0.000--0.000) [$p<0.001$] & \underline{0.000} (0.000--0.000) [$p<0.001$] & \underline{0.000} (0.000--0.000) [$p<0.001$] & \underline{0.000} (0.000--0.000) [$p<0.001$] & - & \underline{0.000} (0.000--0.000) [$p<0.001$] & \underline{0.000} (0.000--0.000) [$p<0.001$] & \textbf{1.000} (1.000--1.000) [-] & \textbf{1.000} (1.000--1.000) [-] & \textbf{1.000} (1.000--1.000) [-] \\
\nolinkurl{CD-A5C-LVOT-REF} & 32 & 0.500 (0.200--0.800) [$p=0.006$] & 0.184 (0.062--0.333) [$p<0.001$] & 0.269 (0.098--0.441) [$p<0.001$] & \underline{0.533} (0.385--0.682) [$p<0.001$] & \underline{0.750} (0.591--0.893) [$p=0.017$] & \underline{0.623} (0.483--0.741) [$p<0.001$] & 0.415 (0.263--0.571) [$p<0.001$] & 0.531 (0.355--0.706) [$p<0.001$] & 0.464 (0.312--0.600) [$p<0.001$] & \textbf{0.971} (0.895--1.000) [-] & \textbf{0.971} (0.895--1.000) [-] & \textbf{0.971} (0.915--1.000) [-] \\
\nolinkurl{CD-PLAX-RVOT} & 25 & 0.000 (0.000--0.000) [$p<0.001$] & 0.000 (0.000--0.000) [$p<0.001$] & 0.000 (0.000--0.000) [$p<0.001$] & \underline{0.500} (0.222--0.778) [$p=0.002$] & \underline{0.278} (0.107--0.471) [$p<0.001$] & \underline{0.356} (0.150--0.545) [$p<0.001$] & 0.400 (0.000--1.000) [$p=0.085$] & 0.074 (0.000--0.207) [$p<0.001$] & 0.125 (0.000--0.308) [$p<0.001$] & \textbf{1.000} (1.000--1.000) [-] & \textbf{0.926} (0.800--1.000) [-] & \textbf{0.962} (0.889--1.000) [-] \\
\nolinkurl{CD-PLAX-REF} & 48 & 0.312 (0.229--0.400) [$p<0.001$] & \underline{0.750} (0.622--0.868) [$p<0.001$] & 0.440 (0.342--0.533) [$p<0.001$] & \underline{0.750} (0.615--0.873) [$p<0.001$] & 0.689 (0.551--0.816) [$p<0.001$] & \underline{0.717} (0.602--0.814) [$p<0.001$] & 0.479 (0.364--0.595) [$p<0.001$] & 0.730 (0.596--0.851) [$p<0.001$] & 0.577 (0.466--0.678) [$p<0.001$] & \textbf{1.000} (1.000--1.000) [-] & \textbf{0.980} (0.930--1.000) [-] & \textbf{0.990} (0.964--1.000) [-] \\
\nolinkurl{CD-PLAX-TV} & 23 & 0.000 (0.000--0.000) [$p<0.001$] & 0.000 (0.000--0.000) [$p<0.001$] & 0.000 (0.000--0.000) [$p<0.001$] & \underline{0.625} (0.421--0.818) [$p<0.001$] & \underline{0.652} (0.444--0.844) [$p=0.067$] & \underline{0.638} (0.455--0.784) [$p<0.001$] & 0.391 (0.212--0.581) [$p<0.001$] & 0.478 (0.269--0.688) [$p=0.002$] & 0.429 (0.245--0.593) [$p<0.001$] & \textbf{0.957} (0.850--1.000) [-] & \textbf{0.917} (0.778--1.000) [-] & \textbf{0.936} (0.842--1.000) [-] \\
\nolinkurl{CD-PSAX-BPA} & 7 & - & \underline{0.000} (0.000--0.000) [$p=0.735$] & \underline{0.000} (0.000--0.000) [$p=0.735$] & - & \underline{0.000} (0.000--0.000) [$p=0.735$] & \underline{0.000} (0.000--0.000) [$p=0.735$] & - & \underline{0.000} (0.000--0.000) [$p=0.735$] & \underline{0.000} (0.000--0.000) [$p=0.735$] & \textbf{0.333} (0.000--1.000) [-] & \textbf{0.125} (0.000--0.500) [-] & \textbf{0.182} (0.000--0.545) [-] \\
\nolinkurl{CD-PSAX-AV-COR} & 22 & 0.151 (0.084--0.225) [$p<0.001$] & 0.684 (0.474--0.875) [$p=0.002$] & 0.246 (0.145--0.348) [$p<0.001$] & 0.568 (0.405--0.727) [$p=0.039$] & \underline{0.957} (0.846--1.000) [$p=0.735$] & \underline{0.712} (0.564--0.831) [$p=0.029$] & \underline{0.654} (0.448--0.846) [$p=0.350$] & 0.684 (0.474--0.875) [$p=0.001$] & 0.667 (0.483--0.810) [$p=0.027$] & \textbf{0.760} (0.593--0.909) [-] & \textbf{1.000} (1.000--1.000) [-] & \textbf{0.864} (0.744--0.952) [-] \\
\nolinkurl{CD-PSAX-MVPM} & 9 & \textbf{1.000} (1.000--1.000) [$p=1.000$] & 0.100 (0.000--0.375) [$p<0.001$] & 0.182 (0.000--0.545) [$p<0.001$] & \textbf{1.000} (1.000--1.000) [$p=1.000$] & \underline{0.800} (0.444--1.000) [$p=0.268$] & \underline{0.889} (0.615--1.000) [$p=0.268$] & \underline{0.833} (0.333--1.000) [$p=0.724$] & 0.444 (0.111--0.800) [$p=0.013$] & 0.571 (0.182--0.857) [$p=0.005$] & \textbf{1.000} (1.000--1.000) [-] & \textbf{1.000} (1.000--1.000) [-] & \textbf{1.000} (1.000--1.000) [-] \\
\nolinkurl{CD-PSAX-RVOT} & 38 & 0.130 (0.000--0.296) [$p<0.001$] & 0.075 (0.000--0.176) [$p<0.001$] & 0.095 (0.000--0.211) [$p<0.001$] & \underline{0.450} (0.324--0.577) [$p<0.001$] & \underline{0.714} (0.559--0.851) [$p=0.005$] & \underline{0.551} (0.424--0.661) [$p<0.001$] & 0.329 (0.225--0.437) [$p<0.001$] & 0.659 (0.500--0.808) [$p<0.001$] & 0.438 (0.318--0.547) [$p<0.001$] & \textbf{0.708} (0.576--0.829) [-] & \textbf{0.950} (0.865--1.000) [-] & \textbf{0.810} (0.709--0.891) [-] \\
\nolinkurl{CD-PSAX-PVein} & 3 & 0.000 (0.000--0.000) [$p=0.734$] & 0.000 (0.000--0.000) [$p=0.667$] & 0.000 (0.000--0.000) [$p=0.734$] & \textbf{0.167} (0.000--0.444) [$p=0.926$] & \textbf{0.667} (0.000--1.000) [$p=0.671$] & \textbf{0.267} (0.000--0.588) [$p=0.789$] & \underline{0.143} (0.000--0.500) [$p=1.000$] & \underline{0.333} (0.000--1.000) [$p=1.000$] & \underline{0.200} (0.000--0.600) [$p=1.000$] & 0.125 (0.000--0.500) [-] & \underline{0.333} (0.000--1.000) [-] & 0.182 (0.000--0.545) [-] \\
\nolinkurl{CD-PSAX-AV-REF} & 23 & 0.000 (0.000--0.000) [$p<0.001$] & 0.000 (0.000--0.000) [$p<0.001$] & 0.000 (0.000--0.000) [$p<0.001$] & 0.615 (0.333--0.889) [$p=0.370$] & \underline{0.346} (0.154--0.556) [$p=0.069$] & 0.440 (0.216--0.634) [$p=0.063$] & \underline{0.733} (0.429--1.000) [$p=0.974$] & \underline{0.346} (0.156--0.556) [$p=0.036$] & \underline{0.467} (0.240--0.667) [$p=0.075$] & \textbf{0.739} (0.524--0.929) [-] & \textbf{0.609} (0.400--0.810) [-] & \textbf{0.667} (0.474--0.815) [-] \\
\nolinkurl{CD-PSAX-TV} & 23 & 0.000 (0.000--0.000) [$p<0.001$] & 0.000 (0.000--0.000) [$p<0.001$] & 0.000 (0.000--0.000) [$p<0.001$] & \underline{0.579} (0.350--0.800) [$p=0.003$] & \underline{0.478} (0.269--0.688) [$p=0.009$] & \underline{0.522} (0.316--0.694) [$p<0.001$] & 0.429 (0.000--0.833) [$p=0.035$] & 0.125 (0.000--0.286) [$p<0.001$] & 0.194 (0.000--0.390) [$p<0.001$] & \textbf{0.952} (0.833--1.000) [-] & \textbf{0.833} (0.652--0.962) [-] & \textbf{0.885} (0.762--0.973) [-] \\
\nolinkurl{CD-SC-SAG-DAo} & 21 & 0.667 (0.000--1.000) [$p=0.673$] & 0.091 (0.000--0.240) [$p<0.001$] & 0.154 (0.000--0.378) [$p<0.001$] & \underline{0.842} (0.640--1.000) [$p=0.098$] & \underline{0.719} (0.500--0.900) [$p=0.338$] & \underline{0.769} (0.593--0.900) [$p=0.072$] & 0.455 (0.143--0.778) [$p=0.005$] & 0.235 (0.059--0.438) [$p<0.001$] & 0.308 (0.087--0.514) [$p<0.001$] & \textbf{1.000} (1.000--1.000) [-] & \textbf{0.864} (0.688--1.000) [-] & \textbf{0.927} (0.815--1.000) [-] \\
\nolinkurl{CD-SC-SAG-IVC} & 16 & 0.000 (0.000--0.000) [$p<0.001$] & 0.000 (0.000--0.000) [$p<0.001$] & 0.000 (0.000--0.000) [$p<0.001$] & \underline{0.609} (0.400--0.810) [$p=0.032$] & \underline{0.882} (0.688--1.000) [$p=0.273$] & \underline{0.718} (0.526--0.863) [$p=0.014$] & 0.341 (0.194--0.500) [$p<0.001$] & 0.818 (0.600--1.000) [$p=0.098$] & 0.480 (0.300--0.635) [$p<0.001$] & \textbf{0.850} (0.667--1.000) [-] & \textbf{1.000} (1.000--1.000) [-] & \textbf{0.919} (0.800--1.000) [-] \\
\nolinkurl{CD-SC-SAG-SAX-AS} & 1 & - & \textbf{0.000} (0.000--0.000) [$p=1.000$] & \textbf{0.000} (0.000--0.000) [$p=1.000$] & - & \textbf{0.000} (0.000--0.000) [$p=1.000$] & \textbf{0.000} (0.000--0.000) [$p=1.000$] & - & \textbf{0.000} (0.000--0.000) [$p=1.000$] & \textbf{0.000} (0.000--0.000) [$p=1.000$] & \textbf{0.000} (0.000--0.000) [-] & \textbf{0.000} (0.000--0.000) [-] & \textbf{0.000} (0.000--0.000) [-] \\
\nolinkurl{CD-SC-SAG-SAX-REF} & 13 & 0.500 (0.000--1.000) [$p=0.589$] & 0.071 (0.000--0.250) [$p<0.001$] & 0.125 (0.000--0.400) [$p<0.001$] & \underline{0.571} (0.300--0.833) [$p=0.079$] & \underline{0.615} (0.333--0.889) [$p=0.249$] & \underline{0.593} (0.333--0.789) [$p=0.065$] & 0.375 (0.000--0.750) [$p=0.021$] & 0.222 (0.000--0.500) [$p<0.001$] & 0.273 (0.000--0.533) [$p<0.001$] & \textbf{0.800} (0.545--1.000) [-] & \textbf{0.857} (0.615--1.000) [-] & \textbf{0.818} (0.609--0.957) [-] \\
\nolinkurl{CD-SC-SAG-SAX-SVC} & 3 & - & \textbf{0.000} (0.000--0.000) [$p=1.000$] & \textbf{0.000} (0.000--0.000) [$p=1.000$] & - & \textbf{0.000} (0.000--0.000) [$p=1.000$] & \textbf{0.000} (0.000--0.000) [$p=1.000$] & - & \textbf{0.000} (0.000--0.000) [$p=1.000$] & \textbf{0.000} (0.000--0.000) [$p=1.000$] & - & \textbf{0.000} (0.000--0.000) [-] & \textbf{0.000} (0.000--0.000) [-] \\
\nolinkurl{CD-SS-LAX-ARCH} & 23 & 0.100 (0.000--0.229) [$p<0.001$] & 0.125 (0.000--0.286) [$p<0.001$] & 0.111 (0.000--0.241) [$p<0.001$] & \underline{0.500} (0.333--0.667) [$p<0.001$] & \underline{0.786} (0.600--0.947) [$p=0.247$] & \underline{0.610} (0.444--0.745) [$p<0.001$] & 0.485 (0.312--0.656) [$p<0.001$] & 0.700 (0.500--0.880) [$p=0.014$] & 0.571 (0.400--0.714) [$p<0.001$] & \textbf{1.000} (1.000--1.000) [-] & \textbf{0.919} (0.778--1.000) [-] & \textbf{0.958} (0.875--1.000) [-] \\
\nolinkurl{CD-SS-LAX-AAo} & 6 & - & \underline{0.000} (0.000--0.000) [$p<0.001$] & \underline{0.000} (0.000--0.000) [$p<0.001$] & \underline{0.000} (0.000--0.000) [$p=0.003$] & \underline{0.000} (0.000--0.000) [$p<0.001$] & \underline{0.000} (0.000--0.000) [$p=0.002$] & - & \underline{0.000} (0.000--0.000) [$p<0.001$] & \underline{0.000} (0.000--0.000) [$p<0.001$] & \textbf{0.875} (0.500--1.000) [-] & \textbf{1.000} (1.000--1.000) [-] & \textbf{0.933} (0.667--1.000) [-] \\
\nolinkurl{CD-SS-SAX-ARCH} & 1 & - & \textbf{0.000} (0.000--0.000) [$p=1.000$] & \textbf{0.000} (0.000--0.000) [$p=1.000$] & - & \textbf{0.000} (0.000--0.000) [$p=1.000$] & \textbf{0.000} (0.000--0.000) [$p=1.000$] & - & \textbf{0.000} (0.000--0.000) [$p=1.000$] & \textbf{0.000} (0.000--0.000) [$p=1.000$] & - & \textbf{0.000} (0.000--0.000) [-] & \textbf{0.000} (0.000--0.000) [-] \\
\nolinkurl{CD-SS-SAX-PA} & 10 & - & 0.000 (0.000--0.000) [$p=0.098$] & 0.000 (0.000--0.000) [$p=0.098$] & \underline{0.500} (0.000--1.000) [$p=0.720$] & \underline{0.091} (0.000--0.333) [$p=0.266$] & \underline{0.154} (0.000--0.462) [$p=0.212$] & - & 0.000 (0.000--0.000) [$p=0.098$] & 0.000 (0.000--0.000) [$p=0.098$] & \textbf{0.750} (0.000--1.000) [-] & \textbf{0.286} (0.000--0.625) [-] & \textbf{0.421} (0.000--0.727) [-] \\
\nolinkurl{CD-SS-SAX-PVein} & 17 & - & 0.000 (0.000--0.000) [$p<0.001$] & 0.000 (0.000--0.000) [$p<0.001$] & \underline{0.750} (0.400--1.000) [$p=0.771$] & \underline{0.350} (0.125--0.600) [$p=0.017$] & \underline{0.476} (0.200--0.700) [$p=0.031$] & - & 0.000 (0.000--0.000) [$p<0.001$] & 0.000 (0.000--0.000) [$p<0.001$] & \textbf{0.818} (0.600--1.000) [-] & \textbf{0.769} (0.538--0.947) [-] & \textbf{0.788} (0.600--0.923) [-] \\
\nolinkurl{CD-SC-COR-LAX-Ao} & 1 & - & \underline{0.000} (0.000--0.000) [$p<0.001$] & \underline{0.000} (0.000--0.000) [$p<0.001$] & - & \underline{0.000} (0.000--0.000) [$p<0.001$] & \underline{0.000} (0.000--0.000) [$p<0.001$] & - & \underline{0.000} (0.000--0.000) [$p<0.001$] & \underline{0.000} (0.000--0.000) [$p<0.001$] & \textbf{1.000} (1.000--1.000) [-] & \textbf{1.000} (1.000--1.000) [-] & \textbf{1.000} (1.000--1.000) [-] \\
\nolinkurl{CD-SC-COR-LAX-REF} & 11 & 0.250 (0.000--1.000) [$p=0.066$] & 0.083 (0.000--0.300) [$p<0.001$] & 0.125 (0.000--0.400) [$p<0.001$] & \underline{0.833} (0.556--1.000) [$p=0.272$] & 0.833 (0.556--1.000) [$p=0.269$] & \underline{0.824} (0.588--0.963) [$p=0.037$] & 0.476 (0.261--0.696) [$p<0.001$] & \underline{0.917} (0.700--1.000) [$p=0.733$] & 0.625 (0.389--0.800) [$p<0.001$] & \textbf{1.000} (1.000--1.000) [-] & \textbf{1.000} (1.000--1.000) [-] & \textbf{1.000} (1.000--1.000) [-] \\
\textbf{Average} & 978 & 0.261 (0.219--0.306) [$p<0.001$] & 0.274 (0.246--0.302) [$p<0.001$] & 0.195 (0.170--0.221) [$p<0.001$] & \underline{0.622} (0.585--0.659) [$p<0.001$] & \underline{0.609} (0.579--0.640) [$p<0.001$] & \underline{0.579} (0.546--0.612) [$p<0.001$] & 0.488 (0.443--0.531) [$p<0.001$] & 0.464 (0.433--0.495) [$p<0.001$] & 0.393 (0.360--0.427) [$p<0.001$] & \textbf{0.902} (0.882--0.920) [-] & \textbf{0.889} (0.868--0.908) [-] & \textbf{0.884} (0.862--0.904) [-] \\
\end{longtable}
\endgroup
\end{landscape}

\begin{figure}[!htbp]
    \centering
    \includegraphics[width=0.99\textwidth]{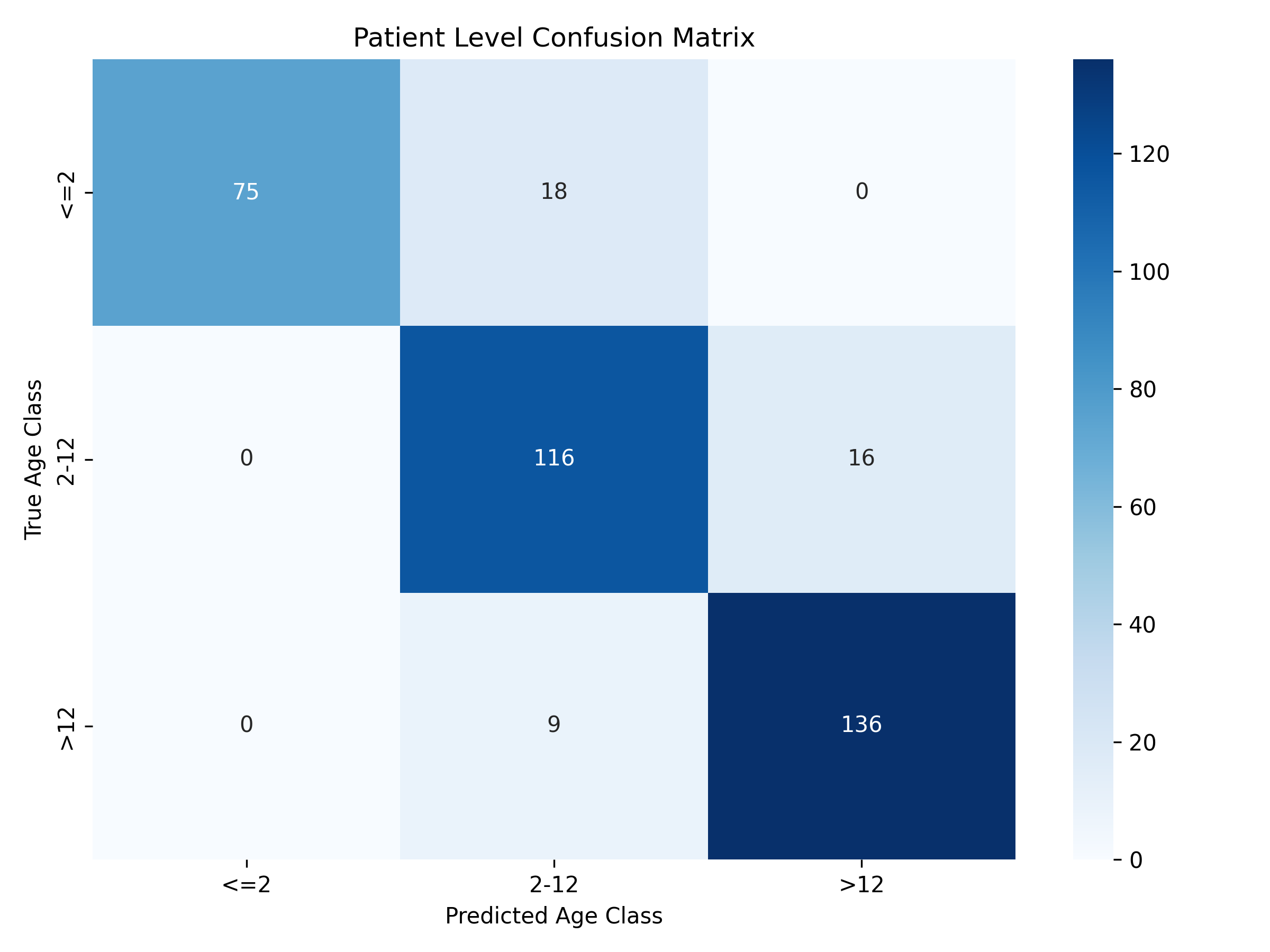}
    \caption{\textbf{Patient-level age-interval classification by EchoDino-PATCH.} Continuous patient-level age estimates were assigned to the intervals $[0,2]$, $(2,12]$, and $(12,18]$ years. Rows indicate chronological-age intervals, columns indicate predicted intervals, and cells show patient counts. Frame-level estimates were aggregated within each patient before interval assignment.}
    \label{si-fig:cm_TCHAge}
\end{figure}

\suppnote{Detailed age-prediction analysis}{si-note:age-prediction}
The confusion matrix (Supplementary Fig.~\ref{si-fig:cm_TCHAge}) summarizes patient-level age-interval predictions in the held-out TCH-Age test set. Continuous age estimates were aggregated across eligible frames for each patient and assigned to the three intervals shown in the matrix. Of 370 patients, 327 (88.4\%) were assigned to the correct interval. All 43 misclassifications occurred between adjacent developmental intervals, with no direct misclassification between the youngest and oldest groups. This pattern indicates that the remaining errors were concentrated near developmental interval boundaries.

\clearpage
\suppnote{Exploratory phenotype-stratified analysis of TCH-SHD predictions}{si-note:shd-phenotypes}
The primary TCH-SHD endpoint was binary study-level classification of the prespecified \texttt{SHD\_any} composite, rather than diagnosis-specific prediction. To examine how this binary prediction behaved across component phenotypes, we calculated, for each non-mutually-exclusive label, the proportion of label-positive studies classified as \texttt{SHD\_any}. This post hoc analysis describes phenotype-stratified recall of the composite classifier; it does not estimate phenotype-specific false-positive rates, discrimination, or diagnostic performance. Label support ranged from $n=1$ to $n=202$ studies, so estimates for rare phenotypes should be interpreted as exploratory.

\begin{table}[!htbp]
    \caption{\textbf{Study-level SHD-positive detection rate stratified by detailed TCH-SHD labels.} Video-level logits were averaged within each study before binary SHD\_any classification. The value for each label is the fraction of label-positive studies classified as SHD\_any; Labels are non-mutually exclusive, and $n$ denotes the number of studies carrying each label. Best values are bold, and second-best values are underlined.}
    \label{si-tab:tch_chd_disease_details}
    \centering
    \scriptsize
    \setlength{\tabcolsep}{1pt}
    \renewcommand{\arraystretch}{1.04}
    \begin{tabular}{lccccc}
    \toprule
    \bf Metadata label & \bf $n$ & \bf EchoPrime & \bf PanEcho & \bf DINOv3 & \bf EchoDino \\
    \midrule
    \texttt{D-looped transposition of the great arteries} & 17 & 0.706 & \textbf{1.000} & \underline{0.941} & \textbf{1.000} \\
    \texttt{Common atrioventricular septal defect} & 1 & \underline{0.000} & \textbf{1.000} & \underline{0.000} & \textbf{1.000} \\
    \texttt{Tricuspid atresia} & 17 & \underline{0.824} & \textbf{1.000} & \textbf{1.000} & \textbf{1.000} \\
    \texttt{Truncus arteriosus} & 8 & \underline{0.250} & \textbf{1.000} & \textbf{1.000} & \textbf{1.000} \\
    \texttt{Total anomalous pulmonary venous connection} & 5 & \underline{0.600} & \textbf{0.800} & \textbf{0.800} & \textbf{0.800} \\
    \texttt{Double inlet left ventricle
    } & 12 & 0.417 & \textbf{1.000} & \underline{0.917} & \textbf{1.000} \\
    \texttt{Double outlet left ventricle} & 3 & \textbf{1.000} & \textbf{1.000} & \textbf{1.000} & \textbf{1.000} \\
    \texttt{Ebstein anomaly} & 7 & 0.571 & \textbf{1.000} & \underline{0.857} & \textbf{1.000} \\
    \texttt{Hypoplastic left heart} & 22 & \underline{0.636} & \textbf{1.000} & \textbf{1.000} & \textbf{1.000} \\
    \texttt{Interrupted aortic arch} & 6 & 0.333 & \underline{0.833} & \underline{0.833} & \textbf{1.000} \\
    \texttt{Pulmonary atresia} & 3 & \underline{0.333} & \textbf{1.000} & \textbf{1.000} & \textbf{1.000} \\
    \texttt{Any atrioventricular septal defect} & 34 & 0.500 & \textbf{1.000} & \underline{0.794} & \textbf{1.000} \\
    \texttt{Tetralogy of Fallot} & 41 & 0.390 & \underline{0.976} & \underline{0.976} & \textbf{1.000} \\
    \texttt{Double outlet right ventricle} & 16 & 0.438 & \textbf{1.000} & \underline{0.875} & \textbf{1.000} \\
    \texttt{Atrial septal defect} & 178 & 0.404 & \underline{0.742} & 0.669 & \textbf{0.781} \\
    \texttt{Ventricular septal defect} & 202 & 0.381 & \underline{0.733} & 0.663 & \textbf{0.856} \\
    \texttt{Patent ductus arteriosus} & 100 & 0.510 & \underline{0.730} & 0.690 & \textbf{0.790} \\
    \texttt{L-looped transposition of the great arteries} & 15 & 0.600 & \underline{0.933} & 0.867 & \textbf{1.000} \\
    \texttt{Any aortic coarctation } & 19 & 0.421 & \textbf{0.684} & \underline{0.632} & \textbf{0.684} \\
    \texttt{Moderate-severe mitral regurgitation} & 111 & 0.198 & \underline{0.459} & 0.396 & \textbf{0.505} \\
    \texttt{Moderate-severe tricuspid regurgitation} & 155 & 0.265 & \underline{0.542} & 0.477 & \textbf{0.581} \\
    \texttt{Moderate-severe aortic regurgitation} & 70 & 0.229 & \underline{0.329} & 0.257 & \textbf{0.500} \\
    \texttt{Moderate-severe pulmonary regurgitation} & 74 & 0.176 & \textbf{0.541} & \underline{0.459} & \textbf{0.541} \\
    \texttt{Any cardiomyopathy} & 59 & 0.237 & \underline{0.525} & 0.339 & \textbf{0.542} \\
    \bottomrule
    \end{tabular}
\end{table}

\begin{figure}[!htbp]
    \centering
    \includegraphics[width=0.99\textwidth]
    {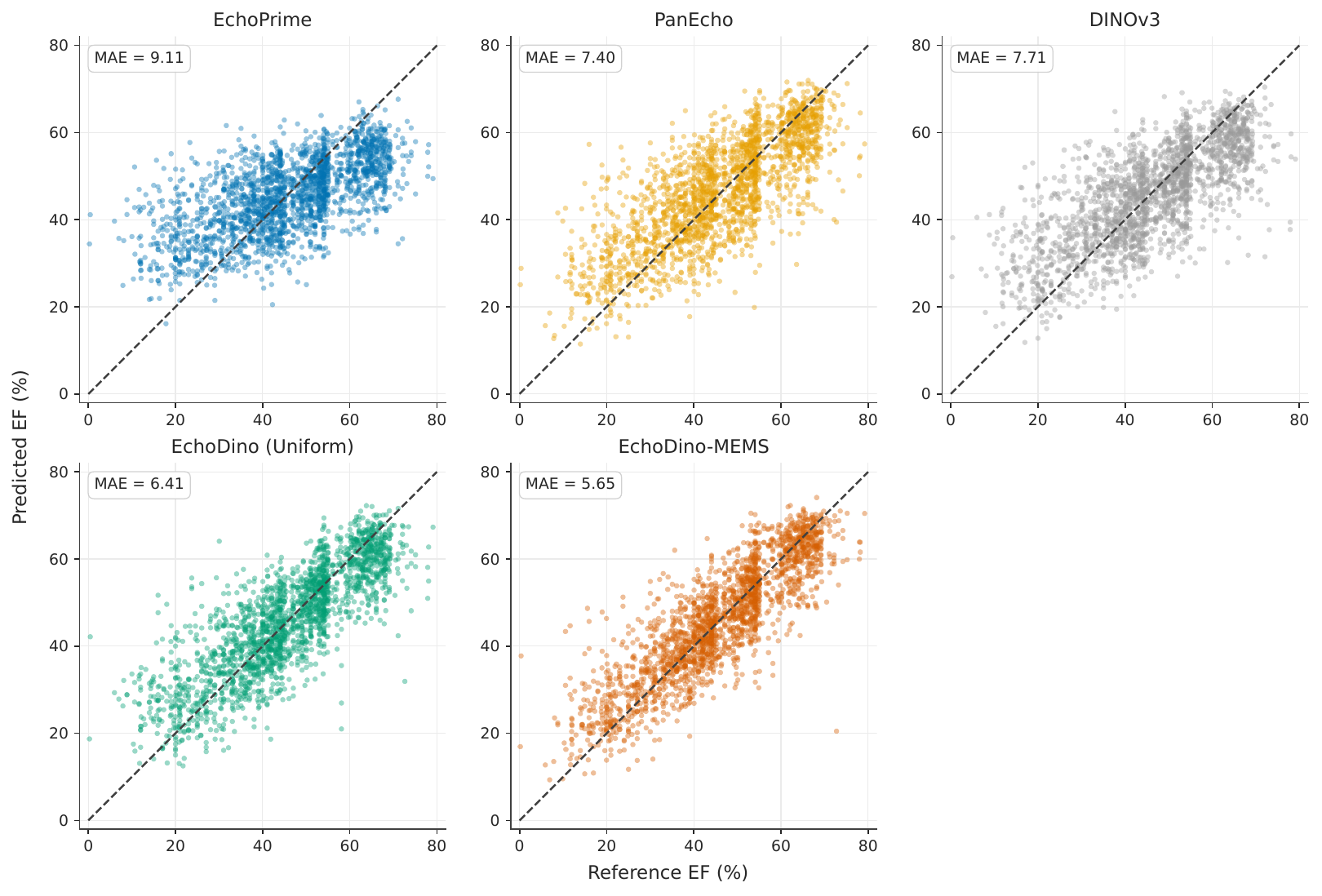}
    \caption{
    \textbf{Predicted versus reference ejection fraction on TCH-EF.}
    Panels compare EchoPrime, PanEcho, DINOv3, EchoDino with uniform
    sampling, and EchoDino-MEMS in reading order. Dashed lines indicate identity ($y=x$). Panel annotations report mean absolute error (MAE) in EF percentage points.
    }
    \label{si-fig:ef-scatter-comparison}
\end{figure}

\clearpage
\suppnote{Prediction-level comparison of ejection-fraction estimates}
{si-note:ef-prediction}
Supplementary Fig.~\ref{si-fig:ef-scatter-comparison} compares predicted
and reference ejection fraction (EF) for the held-out TCH-EF test set.
The shared axes reveal differences in prediction spread and compression
across the reference EF range. EchoDino with uniform sampling achieved
an MAE of 6.41 EF percentage points, which decreased to 5.65 with MEMS,
compared with 9.11 for EchoPrime, 7.40 for PanEcho, and 7.71 for DINOv3.

\suppnote{Ablation study of the proposed MEMS method}{si-note:mems-ablation}

\begin{figure}[!htbp]
    \centering
    \includegraphics[width=0.99\textwidth]{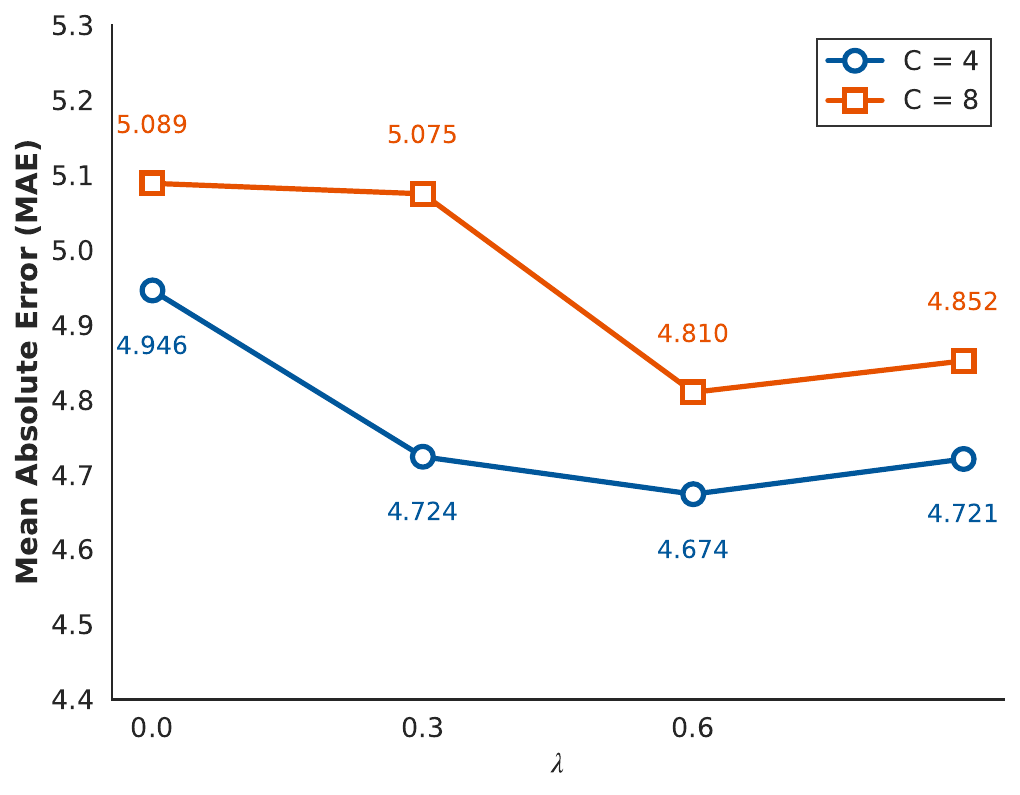}
    \caption{\textbf{Effect of temporal-anchor count and motion bias on ejection-fraction estimation.} Mean absolute error (MAE; EF percentage points) on EchoNet-Dynamic is shown for a fixed 16-frame budget. $C$ denotes the number of uniformly spaced temporal anchors, and $\lambda$ controls the motion bias applied during diversity sampling. Both anchor settings achieved their lowest MAE at $\lambda=0.6$, while $C=4$ yielded lower error than $C=8$ at every tested value. The best configuration ($C=4$, $\lambda=0.6$) achieved an MAE of 4.674. Lower values indicate more accurate EF estimates.}
    \label{si-fig:EchoDynamic_ablation}
\end{figure}

\textbf{MEMS balances temporal coverage with selection of functionally informative cardiac phases.} \label{si-sec: MEMS-ablation} We varied the anchor count $C$ and motion-bias weight $\lambda$ for EF prediction on EchoNet-Dynamic while holding the 16-frame budget fixed (Supplementary Fig.~\ref{si-fig:EchoDynamic_ablation}). For $C=4$, the MAE was 4.946, 4.724, 4.674, and 4.721 at $\lambda=0$, 0.3, 0.6, and 0.9, respectively. At $C=8$, the corresponding MAEs were 5.089, 5.075, 4.810, and 4.852. Both settings achieved their lowest MAE at $\lambda=0.6$. The best configuration ($C=4$, $\lambda=0.6$) achieved an MAE of 4.674. This represented a 0.272 (5.5\%) reduction from diversity-only sampling and a 0.136 (2.8\%) reduction from $C=8$ at the same motion weight. EF estimation requires adequate representation of ventricular states across end diastole, contraction, end systole, and relaxation. With $K=16$, using four anchors left twelve frames for adaptive selection, whereas eight anchors left only eight. The consistently lower MAE with $C=4$ is therefore consistent with preserving whole-video coverage while sampling more distinct functional states. Increasing $\lambda$ to 0.9 slightly increased MAE, indicating that stronger motion weighting provided no additional benefit.

\end{document}